\documentclass[final,twoside]{IEEEtran}
\usepackage{} 

\usepackage{ulem}
\usepackage{float}
\usepackage{color}
\usepackage{amssymb,amsmath,amsthm,bm}

\makeatletter
\newcommand{\tpmod}[1]{{\@displayfalse\pmod{#1}}}
\makeatother

\usepackage{graphicx,color}

\graphicspath{{figures-pdf/}}
\usepackage{cite}
\usepackage{url}
\usepackage{diagbox}
\usepackage[aboveskip=1pt]{subcaption}

\usepackage{algorithm}
\usepackage{bbding}
\usepackage{pifont}
\usepackage{algpseudocode}
\usepackage{multirow,bigstrut}
\usepackage{tabularx}
\usepackage{arydshln}
\usepackage{empheq}
\usepackage{datetime}
\usepackage{algorithmicx}
\usepackage{algpseudocode}
\usepackage{amsmath}
\usepackage{booktabs}

\newlength\imagewidth
\newlength\figsep
\usepackage[bookmarks=false]{hyperref}
\hypersetup{
 linktocpage=true, pdfborderstyle={/S/S/W 1}, hyperindex=true, bookmarks=true, bookmarksopen=true, bookmarksnumbered=true,
}
\hypersetup{hidelinks}

\usepackage{etoolbox}

\usepackage{fancyhdr}
\begin{document}

\title{IGroupSS-Mamba: Interval Group Spatial-Spectral Mamba for Hyperspectral Image Classification}

\author{Yan~He,~\IEEEmembership{Student Member,~IEEE,}
	Bing Tu,~\IEEEmembership{Senior Member,~IEEE,}
    Puzhao Jiang,~\IEEEmembership{Student Member,~IEEE,}
    Bo Liu,~\IEEEmembership{Member,~IEEE,}
	Jun Li,~\IEEEmembership{Fellow,~IEEE,} 
    and Antonio Plaza,~\IEEEmembership{Fellow,~IEEE}	
    \thanks{This work was supported in part by the National Natural Science Foundation of China under Grant 62271200; in part by the Startup Foundation for Introducing Talent of Nanjing University of Information Science and Technology (NUIST) under Grant 2023r091; (Corresponding author: Bing Tu.)}
	\thanks{Yan He, Bing Tu, Puzhao Jiang and Bo Liu are with the Institute of Optics and Electronics, the Jiangsu Key Laboratory for Optoelectronic Detection of Atmosphere and Ocean, and the Jiangsu International Joint Laboratory on Meteorological Photonics and Optoelectronic Detection, Nanjing University of Information Science and Technology, Nanjing, Jiangsu 210044, China (e-mail: 975861884@qq.com; tubing@nuist.edu.cn; jiangpuzhao@126.com; bo@nuist.edu.cn.)}
    \thanks{Jun Li is with the Faculty of Computer Science, China University of Geosciences, Wuhan 430074, China (e-mail: lijuncug@cug.edu.cn).}
    \thanks{Antonio Plaza is with the Hyperspectral Computing Laboratory, Department of Technology of Computers and Communications, Escuela Politecnica, University of Extremadura, 10003 C¨¢ceres, Spain (e-mail: aplaza@unex.es).}
}


\markboth{}{}


\maketitle

\begin{abstract}
Hyperspectral image (HSI) classification has garnered substantial attention in remote sensing fields. Recent Mamba architectures built upon the Selective State Space Models (S6) have demonstrated enormous potential in long-range sequence modeling. However, the high dimensionality of hyperspectral data and information redundancy pose challenges to the application of Mamba in HSI classification, suffering from suboptimal performance and computational efficiency. In light of this, this paper investigates a lightweight Interval Group Spatial-Spectral Mamba framework (IGroupSS-Mamba) for HSI classification, which allows for multi-directional and multi-scale global spatial-spectral information extraction in a grouping and hierarchical manner. Technically, an Interval Group S6 Mechanism (IGSM) is developed as the core component, which partitions high-dimensional features into multiple non-overlapping groups at intervals, and then integrates a unidirectional S6 for each group with a specific scanning direction to achieve non-redundant sequence modeling. Compared to conventional applying multi-directional scanning to all bands, this grouping strategy leverages the complementary strengths of different scanning directions while decreasing computational costs. To adequately capture the spatial-spectral contextual information, an Interval Group Spatial-Spectral Block (IGSSB) is introduced, in which two IGSM-based spatial and spectral operators are cascaded to characterize the global spatial-spectral relationship along the spatial and spectral dimensions, respectively. IGroupSS-Mamba is constructed as a hierarchical structure stacked by multiple IGSSB blocks, integrating a pixel aggregation-based downsampling strategy for multiscale spatial-spectral semantic learning from shallow to deep stages. Extensive experiments demonstrate that IGroupSS-Mamba significantly outperforms the state-of-the-art methods in classification accuracy and achieves lower model parameters and floating point operations (FLOPs).
\end{abstract}

%
\IEEEpeerreviewmaketitle

\section{Introduction}
\label{sec:intro}
\IEEEPARstart{H}{yperspectral} images (HSI) are represented by hundreds of continuous spectral bands in the electromagnetic spectrum, spanning from the visible spectrum, near-infrared, mid-infrared to far-infrared. Benefiting from the abundant spectral and spatial information, HSI classification constitutes the fundamental research in remote sensing fields, which allows for the precise pixel-by-pixel identification and detection of materials for ground objects. This technique has demonstrated widespread applications in various remote sensing scenarios, such as urban planning \cite{ni2020mineral, siebels2020estimation}, military reconnaissance \cite{ardouin2007demonstration, peyghambari2021hyperspectral}, and precision agriculture \cite{lu2020recent, tu2024ncglf2}.

Traditional research on HSI classification predominantly draws upon hand-crafted descriptors and subspace learning methods, such as support vector machine (SVM) \cite{melgani2004classification}, linear discriminant analysis (LDA) \cite{camps2013advances}, and manifold learning \cite{huang2019dimensionality, lunga2013manifold}. To tackle the challenges of spectral variability and spectral confusion, extensive efforts are dedicated to integrating complementary spatial contextual information for more accurate HSI classification, including extended morphological profiles (EMP) \cite{fauvel2008spectral}, extended multi-attribute profiles (EMAP) \cite{dalla2010extended}, and sparse manifold representations \cite{duan2021semisupervised}. However, these methods generally rely on predefined parameter settings, exhibiting insufficient data fitting and description capabilities.

Benefiting from the powerful nonlinear expressive property and automatic feature extraction capability, deep learning-based architectures progressively demonstrate predominant status in HSI classification. For instance, Hu et al. \cite{hu2015deep} first presented a hierarchical 1-D CNN network to extract the high-level spectral features along the spectral dimension of hyperspectral data. Given the significance of spatial correlation in HSIs, Zhong et al. \cite{zhong2017spectral} constructed a 3-D CNN spectral-spatial residual network (SSRN), which performs joint spectral-spatial feature extraction with several 3-D convolutional kernels. Although CNN-based models achieve encouraging performance compared to traditional approaches, they generally struggle to capture global contextual characteristics due to inherent local receptive fields. Comparatively, Transformer benefits from the powerful long-distance sequence modeling capability based on the attention mechanism, bringing significant paradigms for HSI classification. For example, He et al. \cite{he2019hsi} proposed a bidirectional encoder representation transformer network (HSI-BERT). This architecture captures the correlations between pixels through the self-attention mechanism, which provides the global receptive field and mitigates the restriction of spatial distance. Hong et al. \cite{hong2022spectralformer} developed a flexible Transformer-based SpectralFormer (SF) network, capable of processing both pixel-wise and patch-wise inputs. This architecture effectively captures spectrally local sequence information from neighboring bands of HSI data, yielding groupwise spectral embeddings for classification. To reduce model complexity and computational burden, Zhao et al. \cite{zhao2024hyperspectral} introduced a novel groupwise separable convolutional ViT (GSC-ViT), which incorporates a groupwise separable convolution (GSC) module and a groupwise separable multihead self-attention (GSSA) module. The former partitions HSI data into non-overlapping groups for individual convolution operations, while the latter involves lightweight groupwise and pointwise self-attention mechanisms, significantly reducing the overall parameters quantity. Despite the promising classification results of Transformer-based approaches, they typically suffer from quadratic computational complexity $\mathcal{O}\left(N^{2}\right)$ due to the inherent self-attention mechanism, resulting in inefficient modeling and significant memory demands.
\begin{figure}[tb]
    \centering
    \includegraphics[width=0.45\textwidth]{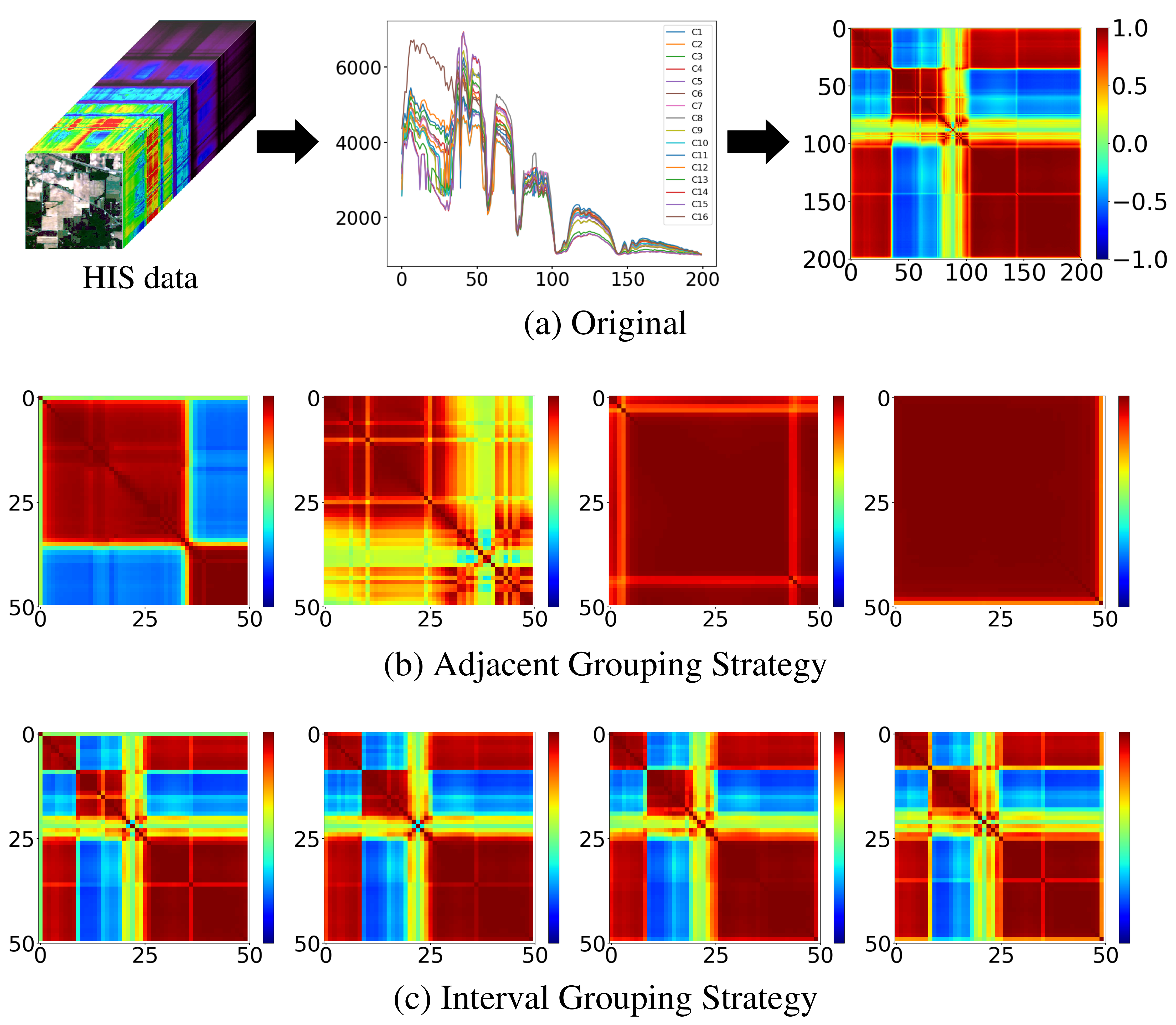}
    \caption{Spectral correlation matrices of Indian Pines dataset. (a) Original spectral band. (b) Spectral group generated by adjacent grouping strategy. (c) Spectral group generated by interval grouping strategy.}
    \label{fig:Correlation}
\end{figure}

Recent Mamba \cite{gu2023mamba} built upon the State Space Models (SSMs) establish long-distance dependency through state transitions, which enjoys the advantages of global contextual modeling, linear computational complexity, and selective information processing. For non-sequential vision data, existing vision Mamba methods \cite{liu2024vmamba} typically extend the Mamba architecture by introducing a multi-directional scanning mechanism. The data scanning is performed horizontally and vertically across spatial dimensions to generate four spatial vision sequences, followed by four independent SSMs for sequence modeling. However, directly applying existing vision Mamba to HSI classification may encounter challenges in achieving satisfactory performance or computational efficiency due to the following limitations:
\begin{itemize}

\item The projection matrices in Mamba are input-dependent and increase linearly with the number of input channels. The high dimensionality of HSIs inevitably imposes substantial computational burdens.

\item Adjacent spectral bands in original HSIs typically exhibit high similarity, as shown in Fig.~\ref{fig:Correlation}(a). Traditional multi-directional scanning strategy applied to all spectral bands may result in information redundancy.

\item Given the abundant spectral information and strong spatial correlation inherent in HSIs, sequence scanning along only the spectral or spatial dimension may lead to the loss of spatial or spectral information, respectively.
\end{itemize}

In response to the aforementioned challenges, this work proposes a lightweight Interval Group Spatial-Spectral Mamba framework (IGroupSS-Mamba) for HSI classification, enabling multi-directional and multi-scale global spatial-spectral information extraction in a grouping and hierarchical manner. Considering the high dimensionality of hyperspectral data and the complementarity of multi-directional scanning of Mamba, an Interval Group S6 Mechanism (IGSM) is introduced. The input high-dimensional features are divided into four non-overlapping groups at intervals along the spectral dimension, with each group followed by a unidirectional S6 for sequence dependency modeling. After the processing, the generated grouping features are concatenated through a channel attention mechanism, addressing the limited interaction inherent in the grouping operation. Following the classic Mamba architecture, IGSM as the fundamental unit is integrated with conventional mapping operations to construct two independent spatial and spectral operators, which are sequentially cascaded and combined with a Feed-Forward Network (FFN) to form the Interval Group Spatial-Spectral Block (IGSSB) for global spatial-spectral feature extraction. Within the spatial and spectral operators, the scanning direction of each group is selected from one of four directions: left-to-right, right-to-left, top-to-bottom, and bottom-to-top, along the spatial and spectral dimensions, respectively. IGroupSS-Mamba is constructed as a hierarchical structure comprising multiple stacked IGSSB blocks, integrated with a pixel aggregation-based downsampling strategy to facilitate the model's multiscale characteristics.
The main contributions are summarized as follows.
\begin{itemize}
\item An Interval Group Spatial-Spectral Mamba framework (IGroupSS-Mamba) based on group modulation and hierarchical mechanism is proposed for HSI classification, which can achieve multi-directional and multi-scale global spatial-spectral information extraction with greater computational efficiency.

\item An Interval Group S6 Mechanism (IGSM) is introduced to perform interval-wise feature grouping and parallel unidirectional sequence scanning, which can enjoy the complementarity of different scanning directions while mitigating information redundancy.

\item An Interval Group Spatial-Spectral Block (IGSSB) is designed. By performing IGSM-based grouping sequence modeling along both spatial and spectral dimensions, the spatial regularity and spectral peculiarity of hyperspectral data can be effectively excavated.

\item Extensive experiments are verified on three public hyperspectral datasets: Indian Pines, Pavia University, and Houston 2013. The results indicate the effectiveness and superiority of the proposed method.

\end{itemize}

The remaining sections of this paper are organized as follows. Sec.~\ref{sec:related} reviewed the related
work. Sec.~\ref{sec:dnn} provides a comprehensive description of the proposed method. Sec.~\ref{sec:simu} outlines the experimental results and analyses. Conclusions and future work are discussed in Sec.~\ref{sec:conc}.

\section{Related work}
\label{sec:related}
\subsection{Hyperspectral Image Classification}
The continuous development of deep learning technology has brought substantial paradigms for HSI classification task. Building upon the properties of local receptive fields and parameter sharing, CNN architectures demonstrate progressive innovations. For instance, Hu et al. \cite{hu2015deep} first proposed a hierarchical 1-D CNN network, which captures the high-level spectral features of hyperspectral data along the spectral dimension. Given the characteristics of abundant spectral channels and strong spatial correlation in HSIs, Lee et al. \cite{lee2016contextual} introduced a contextual deep 2-D CNN to jointly leverage both spatial and spectral features for HSI classification. In contrast to 2-D CNN extracting features by separating the spatial and spectral dimensions, 3-D convolutional kernels possess the capability for joint spectral-spatial feature extraction. Classically, Hamida et al. \cite{hamida20183} developed a 3-D CNN architecture, which captures finer spectral-spatial representations with several stacked 3-D convolutional kernels. To effectively reduce computational complexity, Roy et al. \cite{roy2020hybridsn} constructed a hybrid spectral architecture (HybridSN) that stacks multiple spectral-spatial 3-D CNN blocks and spatial 2-D CNN blocks. The 3-D CNN enables the joint spatial-spectral feature representation from spectral bands, while the 2-D CNN in higher layers captures abstract-level spatial features. As the mainstream backbone architecture, CNN-based methods exhibit formidable capability in local feature extraction; however, they generally encounter challenges in establishing global dependencies between pixels.

Benefiting from the long-range sequence modeling capability based on attention mechanism, Transformer architectures have been significantly investigated for HSI classification task \cite{yang2022hyperspectral, roy2023spectral, zhang2022convolution, 10038735}. In particular, He et al. \cite{he2019hsi} proposed a bidirectional encoder representation transformer network (HSI-BERT), primarily built on a multihead self-attention (MHSA) mechanism in an MHSA layer. Hong et al. \cite{hong2022spectralformer} constructed a pure transformer-based SpectralFormer (SF) network, which involves a groupwise spectral embedding (GSE) to learn localized spectral representations and a cross-layer adaptive fusion (CAF) to convey features from shallow to deep layers. To comprehensively leverage both spectral and spatial information, Zhong et al. \cite{zhong2022spectral} developed a novel spectral-spatial transformer network (SSTN), which consists of spatial attention and spectral association modules to break the long-range constraints of convolution kernels. A factorized architecture search model (FAS) is integrated to determine the layer-level choices and block-level orders. Furthermore, Peng et al. \cite{peng2022spatial} devised a dual-branch spatial-spectral transformer with cross-attention (CASST), in which the spectral branch establishes dependencies among spectral bands and the spatial branch captures mine-grained spatial contexts. The interaction between spatial and spectral information is facilitated within each transformer block via a cross-attention mechanism. To relieve the model complexity and feature over-dispersion, Mei et al. \cite{mei2022hyperspectral} introduced a group-aware hierarchical transformer (GAHT), which confines MHSA to the local spatial-spectral relationships by incorporating a grouped pixel embedding (GPE) module. Zhao et al. \cite{zhao2024hyperspectral} presented a novel groupwise separable convolutional vision transformer (GSC-ViT), which integrates a groupwise separable convolution (GSC) module to partition HSI data into non-overlapping groups for distinct convolution operations. A groupwise separable multi-head self-attention (GSSA) module comprising groupwise self-attention (GSA) and pointwise self-attention (PWSA) is stacked for local and global spatial feature extraction. While Transformer architectures have demonstrated remarkable performance in HSI classification, they suffer from quadratic computational complexity due to inherent self-attention mechanism, presenting substantial challenges in terms of modeling efficiency and memory overhead.

\subsection{State Space Models and Mamba}
As a foundational model in control theory, the classical state-space model \cite{kalman1960new} converts higher-order derivatives into state variables, allowing dynamic systems to be represented through only first-order derivatives. Building upon this foundation, recent State Space Models (SSMs) \cite{gu2022efficiently} capitalize on the strengths of continuous state spaces to model complex temporal dynamics, exhibiting significant advantages in long-distance sequence modeling and linear computational complexity. To enhance dynamic modeling capabilities, Mamba \cite{gu2023mamba} introduces a selective SSMs (S6) architecture, integrating time-varying parameters to selectively process relevant information based on input. Additionally, a hardware-aware mechanism is integrated to facilitate efficient recursive computation.

Drawing inspiration from the effectiveness of Mamba models in sequence data processing, recent studies have extended its potential to vision scenarios involving 2D spatial awareness. For example, Zhu et al. \cite{zhu2024vision} proposed a Vision Mamba (ViM) architecture, which performs bidirectional selective scanning in the horizontal direction, transforming the image into ordered sequential data to perceive global information. Liu et al. \cite{liu2024vmamba} introduced a VMamba framework that incorporates a cross-selective scan module. The data scanning is performed horizontally and vertically across spatial dimensions to generate four spatial vision sequences, achieving effective global contextual modeling with linear complexity. Furthermore, Huang et al. \cite{huang2024localmamba} presented a LocalVMamba model based on windowed selective scan. It introduces a scanning methodology that operates within distinct windows and employs dynamic scanning directions across network layers, effectively capturing local dependencies while maintaining a global perspective. Inspired by these advancements, several researchers have extensively incorporated the Mamba architecture into remote sensing applications, such as change detection \cite{chen2024changemamba}, dense prediction \cite{zhao2024rs}, and semantic segmentation \cite{ma2024rs, zhu2024samba}, significantly highlighting the adaptability and effectiveness of Mamba in handling complex remote sensing signals.

\begin{figure*}[!htb]
	\centering
	\includegraphics[width=0.9\textwidth]{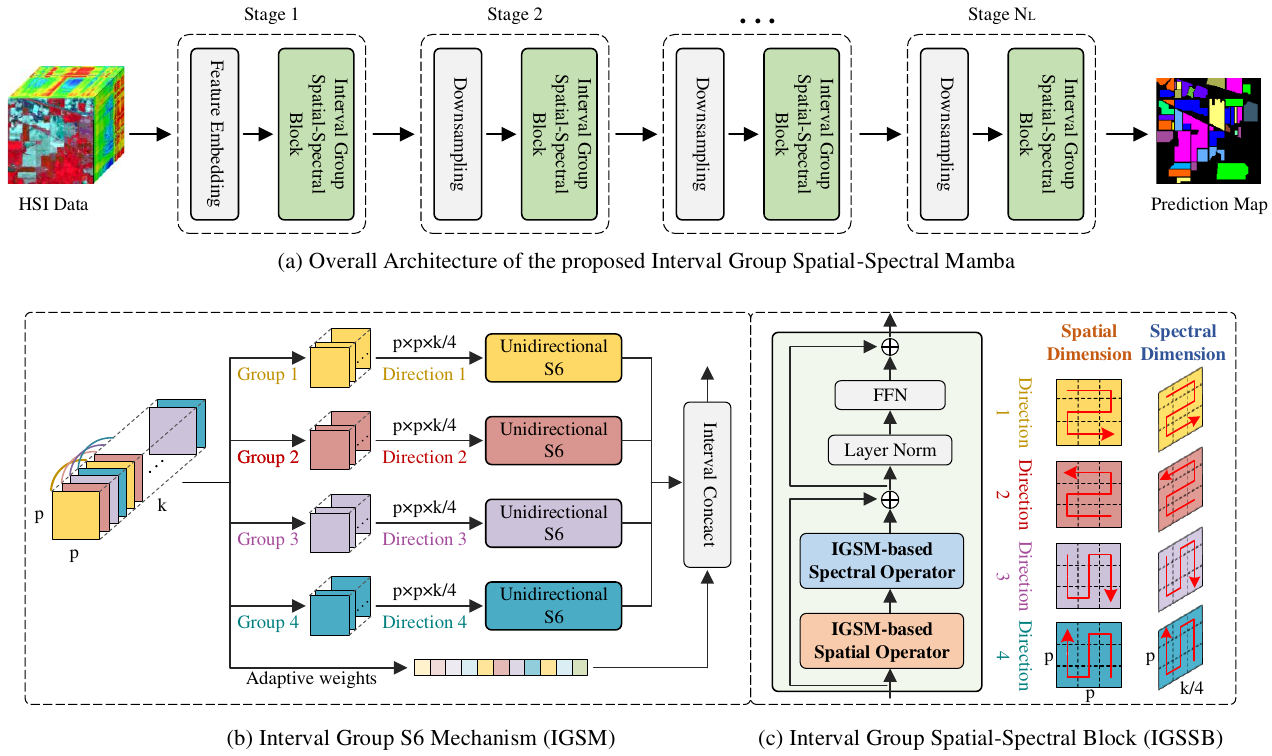}
	\caption{(a) The overall architecture of the proposed Interval Group Spatial-Spectral Mamba framework (IGroupSS-Mamba) for HSI classification; (b) The computational procedure of the proposed Interval Group S6 Mechanism (IGSM); (3) The structural flow of the proposed Interval Group Spatial-Spectral Block (IGSSB).}
\label{fig:Frame}
\end{figure*}

\section{The proposed network}
\label{sec:dnn}
This section commences with the preliminaries associated with State Space Models (SSMs). Following this, we investigate the overall architecture of the proposed IGroupSS-Mamba framework for HSI classification. Subsequently, the sub-model of Interval Group S6 Mechanism (IGSM) is meticulously introduced, followed by the development of Interval Group Spatial-Spectral Block (IGSSB).

\subsection{Preliminaries}
\noindent{\textbf{State Space Models (SSMs).}}
Drawing from the principles of the linear time-invariant system, SSMs are designed to map a one-dimensional signal $x(t) \in \mathbb{R}$ into an output sequence $y(t) \in \mathbb{R}$ via an intermediate hidden state $h(t) \in \mathbb{R}^{N}$. This transformation can be mathematically described through the following linear ordinary differential equation (ODE),
\begin{equation}\label{}
  \begin{aligned}
    h'\left( t \right) &= {\bf{A}}h\left( t \right) + {\bf{B}}x\left( t \right), \\
    y\left( t \right) &= {\bf{C}}h\left( t \right),
  \end{aligned}
\label{eq:hy}
\end{equation}
where $\mathbf{A} \in \mathbb{R}^{N \times N}$ denotes the state transition parameter, and $\mathbf{B} \in \mathbb{R}^{N \times 1}, \mathbf{C} \in \mathbb{R}^{N \times 1}$ represent the projection matrices.

To integrate the continuous-time system depicted in Eq.~(\ref{eq:hy}) into discrete sequence-based deep models, the continuous parameters $\mathbf{A}$ and $\mathbf{B}$ are subsequently discretized via a zero-order hold (ZOH) technique with a time scale parameter $\Delta$. This process can be expressed as

\begin{equation}\label{}
  \begin{aligned}
    \overline{\mathbf{A}} & =\exp (\Delta \mathbf{A}), \\
    \overline{\mathbf{B}} & =(\Delta \mathbf{A})^{-1}(\exp (\Delta \mathbf{A})-\mathbf{I}) \cdot \Delta \mathbf{B} \\
                          & \approx(\Delta \mathbf{A})^{-1}(\Delta \mathbf{A})(\Delta \mathbf{B}) \\
                          & =\Delta \mathbf{B},
  \end{aligned}
\label{eq:AB}
\end{equation}
where $\overline{\mathbf{A}}$ and $\overline{\mathbf{B}}$ represent the discretized forms of the parameters $\bf{A}$ and $\bf{B}$, respectively.

After the discretization step, the discretized SSM system can be formulated as follows
\begin{equation}\label{}
  \begin{aligned}
    h_{t} & =\overline{\mathbf{A}} h_{t-1}+\overline{\mathbf{B}} x_{t}, \\
    y_{t} & =\mathbf{C} h_{t},
  \end{aligned}
\label{eq:SSM}
\end{equation}

For efficient implementation, the aforementioned linear recurrence process is achieved through the following convolution operation, which can be expressed as

\begin{equation}\label{}
  \begin{aligned}
    \overline{\mathbf{K}} & =\left(\mathbf{C} \overline{\mathbf{B}}, \mathbf{C} \overline{\mathbf{A B}}, \ldots, \mathbf{C \overline{A}}^{\mathbf{L}-1} \overline{\mathbf{B}}\right), \\
    \mathbf{y} & =\mathbf{x} * \overline{\mathbf{K}},
  \end{aligned}
\end{equation}
where $L$ denotes the length of input sequence, and $\overline{\mathbf{K}} \in \mathbb{R}^{L}$ serves as the structured convolutional kernel.

\noindent{\textbf{Selective State Space Models (S6).}}
Traditional SSMs predominantly rely on the simplifying assumption of linear time-invariant, which enjoys the advantage of linear time complexity. However, it generally encounters challenges in capturing contextual relationships due to the static parameterizations. To overcome this limitation, Mamba \cite{gu2023mamba} introduces the Selective State Space Models (S6) to enable the interactions between sequential states. Specifically, S6 implements dynamic adjustments to the projection matrices through sequence-aware parameterization, i.e., the parameters $\mathbf{B} \in \mathbb{R}^{B \times L \times N}$, $\mathbf{C} \in \mathbb{R}^{B \times L \times N}$ and $\boldsymbol{\Delta} \in \mathbb{R}^{B \times L \times D}$ are dynamically calculated from the input sequence $x \in \mathbb{R}^{B \times L \times D}$, which facilitates selective perception on each sequence unit.

\subsection{Interval Group Spatial-Spectral Mamba: Overview}
Assuming an original hyperspectral image $I \in {\mathbb{R} ^{H \times W \times V}}$, where $H$ and $W$ represent the spatial dimensions, and $V$ denotes the spectral dimension, principal component analysis (PCA) \cite{renard2008denoising} is initially exploited to alleviate the potential Hughes phenomenon \cite{ma2013hughes}. The modified image is represented as ${I_{PCA}} \in {\mathbb{R}^{H \times W \times L}}$, where $L$ refers to the updated spectral dimension after PCA. Given that neighboring pixels can supplement spatial information for the central pixel, the modified image is further partitioned into a series of 3-D patch cubes $\left\{ {{img} \in {\mathbb{R}^{B \times B \times L}}} \right\}_{i = 1}^{H \times W}$ as input of IGroupSS-Mamba, with labels assigned based on the central pixel of each patch.

The architecture of the proposed Interval Group Spatial-Spectral Mamba (IGroupSS-Mamba) for HSI classification is illustrated in Fig.~\ref{fig:Frame}(a). Adhering to the hierarchical architecture in VMamba \cite{liu2024vmamba}, IGroupSS-Mamba is structured into multiple stages, with each stage comprising a downsampling operation (replaced by pixel embedding in the first stage) and an Interval Group Spatial-Spectral Block (IGSSB).
Initially, the cropped patch undergoes pixel embedding to preserve fine-grained information. Subsequently, the generated shallow features are processed through multiple stacked IGSSB blocks, with downsampling operations integrated to facilitate multiscale semantic representation. Ultimately, the extracted spatial-spectral features are transmitted to the prediction module for final classification.

\subsubsection{Pixel Embedding} This process is implemented with one 3D convolutional block, followed by an embedding operation. Taking the cropped cube $img$ as input, the procedure can be formulated as
\begin{equation}\label{}
  f = {\Phi _{embed}}\left( {{\Phi _{3DConv}}\left( img \right)} \right),
  \label{eq:PE}
\end{equation}
where $f$ denotes the generated shallow spatial-spectral features. The 3D convolution block consists of a 3D convolution layer, a batch normalization layer, and a ReLU activation function. The embedding operation involves a linear layer for dimension transformation.

\subsubsection{Downsampling} The adjacent pixels within a homogeneous region in HSIs generally exhibit similar textural information. Based on this, the downsampling operation is achieved by performing pixel aggregation along the spatial dimension. Given ${f_i} \in \mathbb{R}^{{P_i} \times {P_i} \times {K_i}}$ as the input features in the $i$ $\left\{ {i = 2,3, \ldots {N_L}} \right\}$ stage, the downsampling process can be formulated as
\begin{equation}\label{}
  {f_i} = {\Phi _{embed}}\left( {Aggr\left( {{f_i},\varepsilon  = \left\{ {m \times m,s} \right\}} \right)} \right),
\end{equation}
where $\varepsilon $ signifies the downsampling scale, defined by the aggregation kernel size ${m \times m}$ and the mobility stride $s$. The aggregation operation $Aggr\left(  \cdot  \right)$ is performed through pixel averaging. Similar to Eq.~(\ref{eq:PE}), a linear embedding layer is followed to achieve feature mapping with the dimension of ${d_i}$. Consequently, the corresponding output of the $i$-$th$ stage yields the resolution of ${{p_i} \times {p_i} \times {k_i}}$, where ${p_i} = \left( {\frac{{\left( {{P_i} - m} \right)}}{s} + 1} \right)$ and ${k_i} = {d_i}$.

The downsampling operation adaptively forms multiscale features within the hierarchical architecture, enabling IGroupSS-Mamba to effectively model long-range dependencies with richer contextual semantics and fewer computational costs.

\subsubsection{Interval Group Spatial-Spectral Block} As the core component of IGroupSS-Mamba, the IGSSB is responsible for capturing global spatial-spectral contextual information based on the Interval Group S6 Mechanism (IGSM). After pixel embedding or downsampling in each stage, the mapped features are fed into the IGSSB block. This procedure can be delineated as follows
\begin{equation}\label{}
  f_i^{out} = IGSSB\left( {{f_i}} \right).
\end{equation}
The output feature map $f_i^{out}$ retains the same resolution as the input. More detailed descriptions of IGSM and IGSSB are provided in the following section.

After modeling with ${N_L}$ stages, the extracted spatial-spectral features ${f_{N_L}^{out}}$ undergoes an average pooling operation, and then passes through the $Classifier\left(  \cdot  \right)$ comprised of multilayer perceptron layers to yield the ultimate classification results
\begin{equation}\label{}
  pred = Classifier\left( {{\Phi _{avg}}\left( {f_{N_L}^{out}} \right)} \right).
\end{equation}

\subsection{Interval Group S6 Mechanism}
\label{sec:IGSM}
HSIs are represented by hundreds of continuous spectral bands, typically exhibiting high dimensionality and significant similarity between adjacent bands. Applying traditional multi-directional scanning strategies to all spectral bands will inevitably result in substantial computational demand and information redundancy. To this end,  this paper proposes an Interval Group S6 Mechanism (IGSM), as illustrated in Fig.~\ref{fig:Frame}(b). The key idea is to enable parallel sequence modeling by constructing multiple non-overlapping feature groups, each processed with a specific unidirectional scanning direction. The IGSM primarily consists of three stages: feature interval grouping, unidirectional S6 for sequence modeling, and feature interval concatenation.

\subsubsection{Interval Grouping} Unlike traditional adjacent grouping based on continuous local cubes, the interval grouping strategy takes into account the similarity between adjacent spectral bands. Taking $F = \left\{ {\left[ {{r_1},{r_2}, \ldots {r_k}} \right]} \right\}\left| {{r_i} \in {\mathbb{R}^{p \times p \times 1}}} \right.$ as the input high-dimensional feature, it is partitioned into four non-overlapping groups along the spectral dimension in an interval manner, which can be formulated as
\begin{equation}\label{}
  {G_i} = \left\{ {\left[ {{r_i},{r_{i + 4}}, \ldots {r_{k + i - 4}}} \right]} \right\},i = 1,2,3,4.
  \label{eq:Interval}
\end{equation}
The dimensionality of spectral channels for each group is reduced to one-quarter of the input dimensionality, which enjoys the resolution of  ${G_i} \in \mathbb{R}^{p \times p \times \frac{k}{4}}$. Interval grouping facilitates the construction of non-redundant global information within each group, while reducing data dimensionality to enhance computational efficiency.
\begin{figure}[tb]
    \centering
    \includegraphics[width=0.45\textwidth]{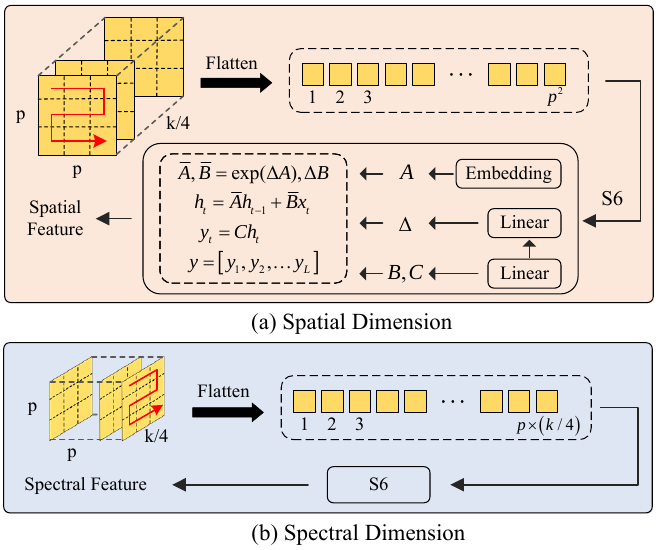}
    \caption{The specific computation procedure for unidirectional S6, taking the scanning direction of left-to-right as an example.}
    \label{fig:S6}
\end{figure}

\subsubsection{Unidirectional S6} To leverage the complementarity of multi-directional scanning, each feature group is followed by a unidirectional S6 for sequence modeling, with the scanning direction selected from one of four routes: left-to-right, right-to-left, top-to-bottom, and bottom-to-top, i.e, ${{v_{LR}}}$, ${{v_{RL}}}$, ${{v_{TB}}}$ and ${{v_{BT}}}$. Considering the intricate spectral correlations and spatial patterns in HSI data, we construct these four directions from both spatial and spectral dimensions, which are respectively applied to the spatial and spectral operators within IGSSB, as depicted in Fig.~\ref{fig:Frame} (c). As a result, taking the grouping features ${G_1}$, ${G_2}$, ${G_3}$, and ${G_4}$ as input, the parallel modeling procedure for each group can be formulated as
\begin{equation}\label{}
  \begin{aligned}
    G_1^{out} &= S6\left( {{G_1},{d_1} = \left\{ {{\upsilon _{LR}},{\omega _{dim}}} \right\}} \right),\\
    G_2^{out} &= S6\left( {{G_2},{d_2} = \left\{ {{\upsilon _{RL}},{\omega _{dim}}} \right\}} \right),\\
    G_3^{out} &= S6\left( {{G_3},{d_3} = \left\{ {{\upsilon _{TB}},{\omega _{dim}}} \right\}} \right),\\
    G_4^{out} &= S6\left( {{G_4},{d_4} = \left\{ {{\upsilon _{BT}},{\omega _{dim}}} \right\}} \right).
  \end{aligned}
\end{equation}
where ${\omega _{dim}}$ refers to either the spatial or spectral dimension. Taking the first group as an example, the specific computation procedure for unidirectional S6 is illustrated in Fig.~\ref{fig:S6}.

Spatial Dimension: The grouping feature ${G_1} \in \mathbb{R}^{p \times p \times \frac{k}{4}}$ first undergoes pixel-wise flattening from left to right along the spatial dimension, generating the flattened 1-D sequence $G_1^{spa} = \left\{ {\left[ {{x_0},{x_1}, \ldots {x_{{p^2}}}} \right]} \right\}\left| {{x_j} \in {\mathbb{R}^{1 \times \frac{k}{4}}}} \right.$. Subsequently, this sequence is transmitted into the S6 model for state space evolution as
\begin{equation}\label{}
  \begin{aligned}
    {h_j} &= {\overline {\bf{A}} _{{\rm{spa}}}}{h_{j - 1}} + {\overline {\bf{B}} _{{\rm{spa}}}}{x_j}\\
    {y_j} &= {{\bf{C}}_{{\rm{spa}}}}{h_j},
  \end{aligned}
\end{equation}
where ${\overline {\bf{A}} _{{\rm{spa}}}}$, ${\overline {\bf{B}} _{{\rm{spa}}}}$, ${{\bf{C}}_{{\rm{spa}}}}$ are the training parameters of the S6 model, and N denotes the hidden state dimension. After efficient computation, the output sequence $G_1^{out} = \left\{ {\left[ {{y_0},{y_1}, \ldots {y_{{p^2}}}} \right]} \right\}\left| {{y_j} \in {\mathbb{R}^{1 \times \frac{k}{4}}}} \right.$ is obtained for subsequent processing. Notably, the same procedure is executed for the remaining three parallel groups, with different scanning directions and independent training parameters.

Spectral Dimension: As illustrated in Fig.~\ref{fig:S6}(b), the grouping feature ${G_1} \in \mathbb{R}^{p \times p\times \frac{k}{4}}$ performs data flattening from left to right along the spectral dimension, yielding the flattened spectral sequence
$G_1^{spe} = \left\{ {\left[ {{x_0},{x_1}, \ldots {x_{p \times \frac{k}{4}}}} \right]} \right\}\left| {{x_j} \in {\mathbb{R}^{1 \times p}}} \right.$. Similarly, this sequence is subsequently transmitted into the S6 model as
\begin{equation}\label{}
  \begin{aligned}
    {h_j} &= {\overline {\bf{A}} _{{\rm{spe}}}}{h_{j - 1}} + {\overline {\bf{B}} _{{\rm{spe}}}}{x_j}\\
    {y_j} &= {{\bf{C}}_{{\rm{spe}}}}{h_j},
  \end{aligned}
\end{equation}
where ${\overline {\bf{A}} _{{\rm{spe}}}}$, ${\overline {\bf{B}} _{{\rm{spe}}}}$, ${{\bf{C}} _{{\rm{spe}}}}$ represents the training parameters. The ultimate transformed output can be expressed as $G_1^{out} = \left\{ {\left[ {{y_0},{y_1}, \ldots {y_{p \times \frac{k}{4}}}} \right]} \right\}\left| {{y_j} \in {\mathbb{R}^{1 \times p}}} \right.$.

\subsubsection{Feature Interval Concatenation} After the parallel group processing, the generated grouping outputs are subsequently subjected to interval concatenation. To encourage inter-group information interaction, dynamic weights are modulated in response to the spectral channels as
\begin{equation}\label{}
  W = \sigma \left( {{\Phi _{atten}}\left( {{\Phi _{AvgPool}}\left( {F} \right)} \right)} \right),
\end{equation}
where ${{\Phi _{AvgPool}}}$ represents the average pooling operation to calculate channel statistics. ${{\Phi _{atten}}}$ consists of two linear layers designed to learn the attention weights for each channel, in conjunction with the non-linearity function $\sigma$. Subsequently, the dynamic weights are leveraged to recalibrate the output grouping features, which can be formulated as
\begin{equation}\label{}
  {F^{out}} = W \cdot InterCat\left( {G_1^{out},G_2^{out},G_3^{out},G_4^{out}} \right).
  \label{eq:out}
\end{equation}
Here, $InterCat\left(  \cdot  \right)$ denotes the interval-wise concatenation operation, which enables each spectral channel to maintain their positional order consistent with the original input.

\begin{figure}[tb]
    \centering
    \includegraphics[width=0.4\textwidth]{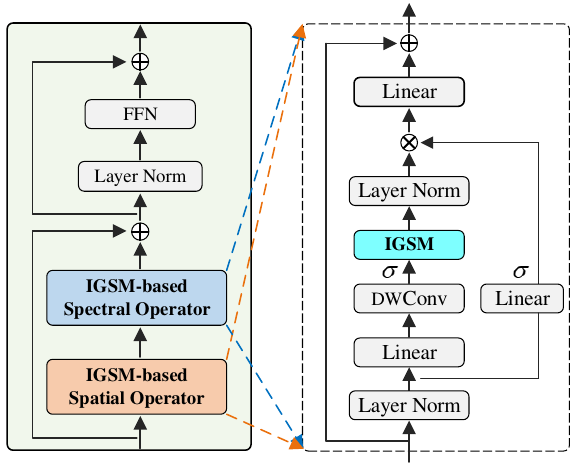}
    \caption{The detailed structure of Interval Group Spatial-Spectral Block (IGSSB).}
    \label{fig:IGSSB}
\end{figure}

\subsection{Interval Group Spatial-Spectral Block}
To effectively exploit the inherent spectral peculiarity and spatial regularity in HSI data, Interval Group Spatial-Spectral Block (IGSSB) integrates both IGSM-based spatial and spectral operators in a cascaded manner, enabling the extraction of global spatial-spectral semantic representations. The detailed structure of IGSSB is illustrated in Fig.~\ref{fig:IGSSB}.

Letting ${f_i}$ represent the input features of the $i$-$th$ stage, the forward process for IGSSB can be mathematically described as
\begin{equation}\label{}
  \begin{aligned}
    {{f'}_i} &= {f_i} + Spe\left( {Spa\left( {{f_i}} \right)} \right) \\
    f_i^{out} &= {{f'}_i} + FFN\left( {{{f'}_i}} \right)
  \end{aligned}
\end{equation}
where $Spa\left(  \cdot  \right)$ and $Spe\left(  \cdot  \right)$ denote the IGSS-based spatial and spectral operators, respectively. The intermediate representation ${{f'}_i}$ is transmitted to the feed-forward network $FFN\left(  \cdot  \right)$ comprising two linear layers and an activation function, which enhances the representation capability of model through nonlinear transformations.

Following the classic Mamba architecture \cite{liu2024vmamba}, the spatial operator $Spa\left(  \cdot  \right)$ and spectral operator $Spe\left(  \cdot  \right)$ perform with consistent forward procedure, as illustrated on the right side of Fig.~\ref{fig:IGSSB}.

Specifically, the operator commences with a normalization layer to enhance the model training stability. Following this, two parallel linear embedding layers are stacked: one branch is followed by an activation function for gating signal generation, while the other branch undergoes a depth-wise convolution (DWConv) operation. The procedure can be formulated as
\begin{equation}\label{}
  z = \sigma \left( {{\Phi _{linear}}\left( {{\Phi _{norm}}\left( f_i \right)} \right)} \right),
\end{equation}
\begin{equation}\label{}
  F = \sigma \left( {{\Phi _{DWConv}}\left( {{\Phi _{linear}}\left( {{\Phi _{norm}}\left( {{f_i}} \right)} \right)} \right)} \right),
\end{equation}
where $\sigma$ denotes the Silu \cite{elfwing2018sigmoid} activation operation. After this, the generated representation passes through the pivotal IGSM layer, executing the group-based sequence scanning as described in Sec.~\ref{sec:IGSM}. The output subsequently undergoes layer normalization and a gating operation. Finally, the features are transmitted to the ultimate linear layer, followed by a residual connection.
\begin{equation}\label{}
  R = {\Phi _{linear}}\left( {{\Phi _{norm}}\left( {IGSM\left( F \right)} \right) \otimes z} \right) + f_i.
\end{equation}
The $IGSM\left(  \cdot  \right)$ layer serves as the core computational unit, with the scanning directions for the spatial and spectral operators illustrated in Fig.~\ref{fig:Frame} (c). Both perform left-to-right, right-to-left, top-to-bottom, and bottom-to-top for the groups, but along the spatial and spectral dimensions, respectively. The output of IGSSB maintains a consistent resolution as input.

\section{Performance evaluation}
\label{sec:simu}

\subsection{Datasets Description}
To validate the classification capabilities of the proposed IGroupSS-Mamba, we conduct comprehensive experimental comparisons on three publicly HSI databases: Indian Pines, Pavia University, and Houston 2013. The division details for training and testing sets are provided in Table~\ref{tab:dataset}.

\subsubsection{Pavia University} The dataset was acquired by the Reflective Optics System Imaging Spectrometer (ROSIS) over Pavia, Northern Italy. The imaging wavelength of the spectrometer ranges from 0.43 to 0.86 $\mu$m. The dataset encompasses 103 spectral bands (after removing 12 noisy bands) and 610 $\times$ 340 pixels, with a spatial resolution of 1.3 m per pixel. There are a total of 42776 ground object pixels, covering 9 distinct categories including Asphalt, Gravel, etc.

\subsubsection{Indian Pines} The dataset was captured by the Airborne/Visible Infrared Imaging Spectrometer (AVIRIS) imaging an Indian pine tree over Northwestern Indiana in 1992. The imaging wavelength of the spectrometer ranges from 0.4 to 2.5 $\mu$m. The dataset consists of 200 spectral bands (after removing the water absorption channels) and 145 $\times$ 145 pixels, with a spatial resolution of 20 m per pixel. There are 10249 ground sample points, representing 16 distinct types including Alfalfa, Corn-notill, etc.

\subsubsection{Houston 2013} The dataset was collected by the ITRES CASI-1500 sensor over the University of Houston campus and its surrounding areas, provided by the 2013 GRSS Data Fusion Contest. The imaging wavelength of the spectrometer ranges from 0.38 to 1.05 $\mu$m. The dataset comprises 144 spectral bands and 340 $\times$ 1905 pixels, with a spatial resolution of 2.5 m per pixel. The total 16373 sample pixels are categorized into 15 challenging types, such as Healthy grass, Trees, etc.

\begin{table*}[!htb]
\centering
\caption{\textsc{Category Name, Sample Numbers of Train Set and Test Set of Each Class on the Indian Pines, Pavia University, and Houston 2013 Datasets.}}
\setlength\tabcolsep{9pt}
\renewcommand\arraystretch{1.1}
\begin{tabular}{cccccccccc}
\hline
                         & \multicolumn{3}{c}{\textbf{Indian Pines}}                        & \multicolumn{3}{c}{\textbf{Pavia University}}                      & \multicolumn{3}{c}{\textbf{Houston 2013}} \\ \hline
\multicolumn{1}{c}{No.} & Category            & Train & \multicolumn{1}{c}{Test} & Category             & Train & \multicolumn{1}{c}{Test}  & Category         & Train & Test  \\ \hline
\multicolumn{1}{c|}{1}   & Alfalfa             & 5     & \multicolumn{1}{c|}{41}   & Asphalt              & 332   & \multicolumn{1}{c|}{6299}  & Healthy grass    & 125   & 1238  \\
\multicolumn{1}{c|}{2}   & Corn-notill         & 143   & \multicolumn{1}{c|}{1285} & Meadows              & 932   & \multicolumn{1}{c|}{17717} & Stressed grass   & 125   & 1241  \\
\multicolumn{1}{c|}{3}   & Corn-mintill        & 83    & \multicolumn{1}{c|}{747}  & Gravel               & 105   & \multicolumn{1}{c|}{1994}  & Synthetic grass  & 70    & 690   \\
\multicolumn{1}{c|}{4}   & Corn                & 24    & \multicolumn{1}{c|}{213}  & Trees                & 153   & \multicolumn{1}{c|}{2911}  & Trees            & 124   & 1231  \\
\multicolumn{1}{c|}{5}   & Grass-pasture & 48    & \multicolumn{1}{c|}{435}  & Painted metal sheets & 67    & \multicolumn{1}{c|}{1278}  & Soil             & 124   & 1229  \\
\multicolumn{1}{c|}{6}   & Grass-trees         & 73    & \multicolumn{1}{c|}{657}  & Bare soil            & 251   & \multicolumn{1}{c|}{4778}  & Water            & 33    & 321   \\
\multicolumn{1}{c|}{7}   & Grass-pasture-mowed       & 3     & \multicolumn{1}{c|}{25}   & Bitumen              & 67    & \multicolumn{1}{c|}{1263}  & Residential      & 127   & 1255  \\
\multicolumn{1}{c|}{8}   & Hay-windrowed       & 48    & \multicolumn{1}{c|}{430}  & Self-blocking bricks & 184   & \multicolumn{1}{c|}{3498}  & Commercial       & 124   & 1231  \\
\multicolumn{1}{c|}{9}   & Oats                & 2     & \multicolumn{1}{c|}{18}   & Shadows              & 47    & \multicolumn{1}{c|}{900}   & Road             & 125   & 1239  \\
\multicolumn{1}{c|}{10}  & Soybean-notill      & 97    & \multicolumn{1}{c|}{875}  &                      &       & \multicolumn{1}{c|}{}      & Highway          & 123   & 1214  \\
\multicolumn{1}{c|}{11}  & Soybean-mintill     & 245   & \multicolumn{1}{c|}{2210} &                      &       & \multicolumn{1}{c|}{}      & Railway          & 123   & 1222  \\
\multicolumn{1}{c|}{12}  & Soybean-clean       & 59    & \multicolumn{1}{c|}{534}  &                      &       & \multicolumn{1}{c|}{}      & Parking Lot 1    & 123   & 1220  \\
\multicolumn{1}{c|}{13}  & Wheat               & 20    & \multicolumn{1}{c|}{185}  &                      &       & \multicolumn{1}{c|}{}      & Parking Lot 2    & 47    & 464   \\
\multicolumn{1}{c|}{14}  & Woods               & 126   & \multicolumn{1}{c|}{1139} &                      &       & \multicolumn{1}{c|}{}      & Tennis Court     & 43    & 423   \\
\multicolumn{1}{c|}{15}  & Buildings           & 39    & \multicolumn{1}{c|}{347}  &                      &       & \multicolumn{1}{c|}{}      & Running Track    & 66    & 653   \\
\multicolumn{1}{c|}{16}  & Stone               & 6     & \multicolumn{1}{c|}{84}   &                      &       & \multicolumn{1}{c|}{}      &                  &       &       \\ \hline
\multicolumn{1}{c|}{}    & Total               & 1024  & \multicolumn{1}{c|}{9225} & Total                & 2138  & \multicolumn{1}{c|}{40638} & Total            & 1502  & 14871 \\
\hline
\end{tabular}
\label{tab:dataset}
\end{table*}

\begin{table*}[!htb]
\centering
\caption{\textsc{ Sensitivity analysis for the proposed method with different sizes of input patches on the Indian Pines, Pavia University, and Houston 2013 Datasets.}}
\setlength\tabcolsep{8.5pt}
\renewcommand\arraystretch{1.1}
\begin{tabular}{c|ccccccccc}
\hline
\multirow{2}{*}{\textbf{Patch Sizes}} & \multicolumn{3}{c}{\textbf{Indian Pines}}                                  & \multicolumn{3}{c}{\textbf{Pavia University}}                                   & \multicolumn{3}{c}{\textbf{Houston2013}}         \\ \cline{2-10}
                                      & OA (\%)       & AA (\%)        & \multicolumn{1}{c|}{Kappa (\%)}     & OA (\%)        & AA (\%)        & \multicolumn{1}{c|}{Kappa (\%)}     & OA (\%)        & AA (\%)        & Kappa (\%)     \\ \hline
$9 \times 9$                                   & 97.44         & 96.42          & \multicolumn{1}{c|}{97.08}          & 99.36          & 99.18          & \multicolumn{1}{c|}{99.15}          & 98.88          & 99.00          & 98.79          \\
$11 \times 11$                                 & 97.18         & 96.33          & \multicolumn{1}{c|}{96.78}          & 99.56          & 99.34          & \multicolumn{1}{c|}{99.42}          & 99.13          & 99.22          & 99.06          \\
$13 \times 13$                                & \textbf{98.7} & \textbf{98.46} & \multicolumn{1}{c|}{\textbf{98.52}} & \textbf{99.75} & \textbf{99.57} & \multicolumn{1}{c|}{\textbf{99.67}} & 99.34          & \textbf{99.39} & 99.29          \\
$15 \times 15$                                 & 98.65         & 98.20          & \multicolumn{1}{c|}{98.46}          & 99.37          & 99.19          & \multicolumn{1}{c|}{99.16}          & \textbf{99.35} & \textbf{99.39} & \textbf{99.30} \\
$17 \times 17$                                 & 97.40         & 96.66          & \multicolumn{1}{c|}{97.03}          & 99.69          & 99.32          & \multicolumn{1}{c|}{99.59}          & 99.05          & 99.1           & 98.97          \\ \hline
\end{tabular}
\label{tab:Patchsize}
\end{table*}

\subsection{Experimental Settings}
\subsubsection{Evaluation Metrics}
Following existing HSI classification approaches, overall classification accuracy (OA), average classification accuracy (AA), and kappa coefficient (Kappa) are exploited as the evaluation metrics. All experiments are performed under identical experimental conditions, with the reported results averaged over five consecutive trials.

\subsubsection{Implementation Details}
All experiments are conducted on the PyTorch platform, utilizing a single RTX 3070Ti GPU. For parameter optimization, the Adam optimizer is employed with a learning rate set at 0.001. The training process is configured with 100 epochs and a batch size of 64. The PCA dimension for reduction is set to 30. Consistent with the default hyperparameters outlined in VMamba \cite{liu2024vmamba}, the state dimension and expansion ratio in S6 mechanism are fixed at 16 and 1, respectively.

\subsubsection{Competitive Approaches}
To demonstrate the effectiveness of the proposed IGroupSS-Mamba, three kinds of representative HSI classification architectures are selected for comprehensive comparison, including conventional methods (SVM \cite{melgani2004classification}), CNN-based methods (2D-CNN, 3D-CNN \cite{hamida20183}, and Transformer-based methods (HSI-BERT \cite{he2019hsi}, SF \cite{hong2022spectralformer}, SSTN \cite{zhong2022spectral}, CASST \cite{peng2022spatial}, GSC-ViT \cite{zhao2024hyperspectral}, SS-MTr \cite{huang2023spectral}).

\subsection{Parameter Analysis}
In this section, a series of experiments are carried out to determine the optimal parameters for IGroupSS-Mamba, including the input patch sizes, the downsampling scales, the embedding dims, and the stacked depths.

\subsubsection{Different Input Patch Sizes}
Table~\ref{tab:Patchsize} presents the impact of different input patch sizes on classification performance, ranging from $9 \times 9$ to $17 \times 17$ with a growth interval of 2. As can be observed, the Houston 2013 dataset displays a stable tendency of consistent increase followed by a decrease, and the Pavia University and Indian Pines datasets exhibit slight fluctuation, with their maximum peak at the identified point $13 \times 13$. Smaller patches may result in insufficient information, whereas larger patches can introduce redundant noise factors, leading to information aliasing. Consequently, a patch size of $13 \times 13$ is leveraged for all datasets in our experiments.

\subsubsection{Different Embedding Dims and Depths}
\begin{figure*}[htbp]
	\centering
    \begin{subfigure}{0.32\linewidth}
		\centering
		\includegraphics[width=5.5cm,height=4cm]{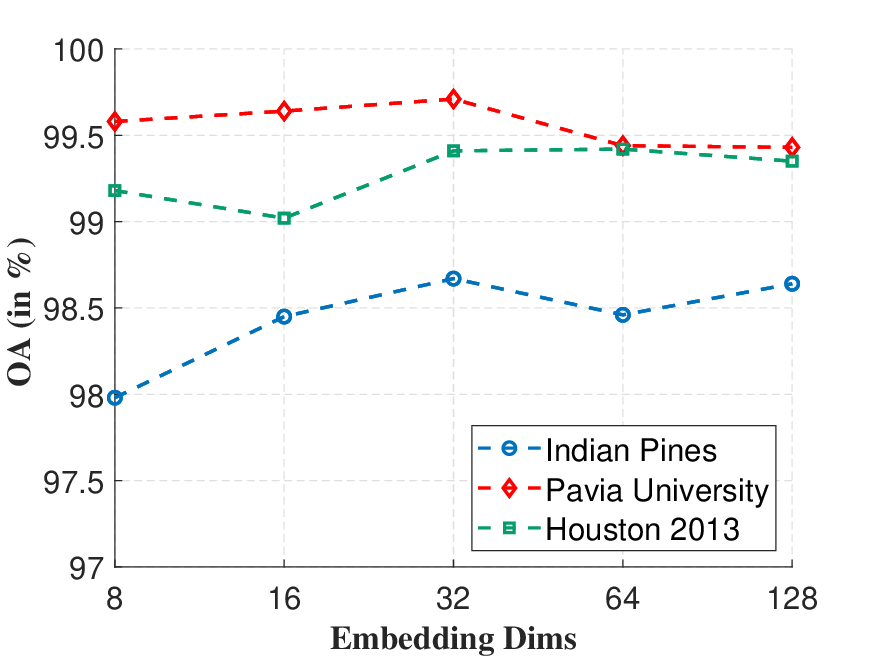}
		\caption{}
        \setlength{\belowdisplayskip}{43pt}
		\label{pseudo-I}
	\end{subfigure}
	\centering
	\begin{subfigure}{0.32\linewidth}
		\centering
		\includegraphics[width=5.5cm,height=4cm]{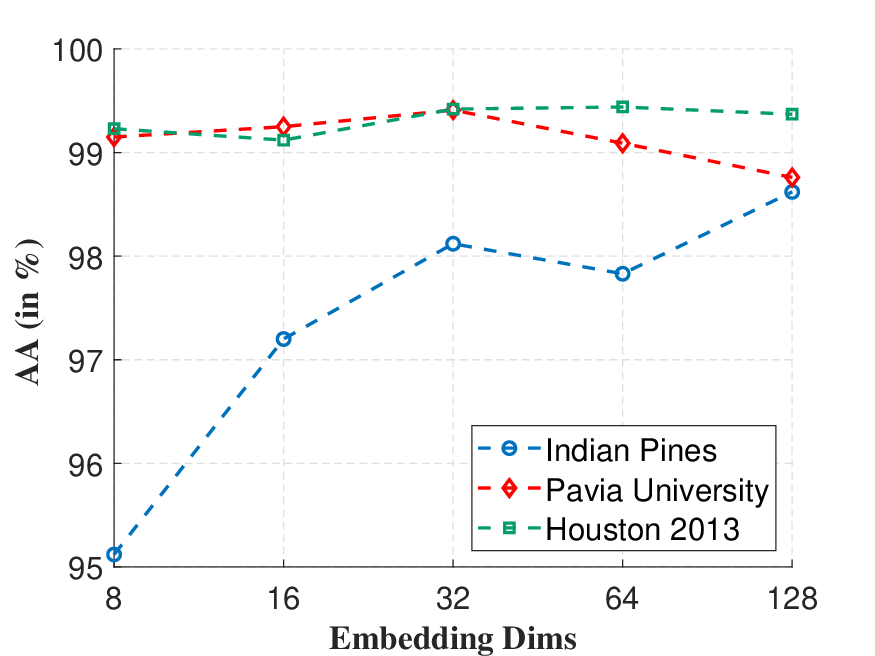}
		\caption{}
		\label{pseudo-I}
	\end{subfigure}
    \begin{subfigure}{0.3\linewidth}
		\centering
		\includegraphics[width=5.5cm,height=4cm]{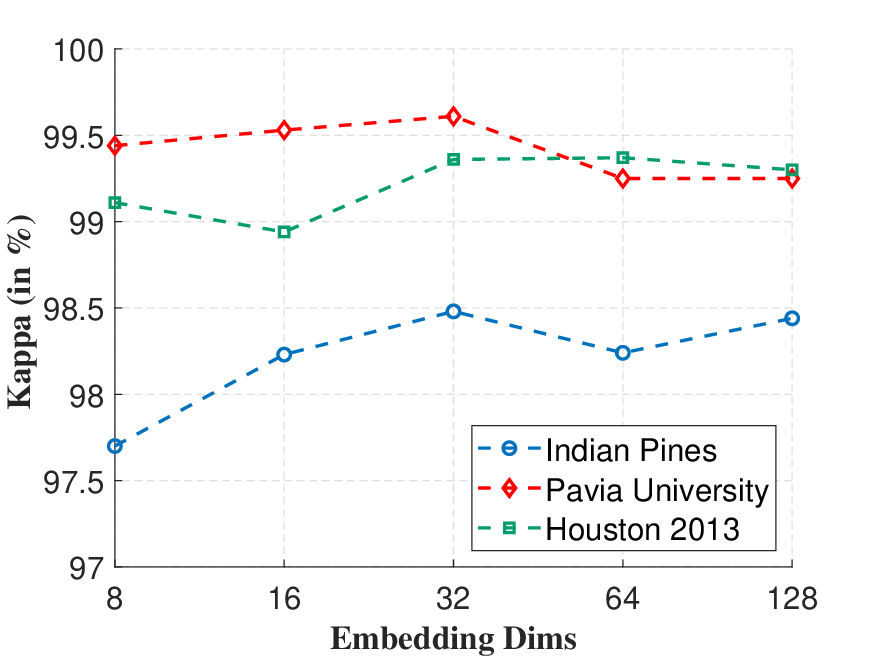}
		\caption{}
		\label{pseudo-I}
	\end{subfigure}
    \caption{Sensitivity analysis for the proposed method with different feature embedding dimensions in terms of (a) OA, (b) AA, and (c) Kappa.}
	\label{fig:Embed}
\end{figure*}

\begin{figure*}[htbp]
	\centering
    \begin{subfigure}{0.32\linewidth}
		\centering
		\includegraphics[width=5.5cm,height=4cm]{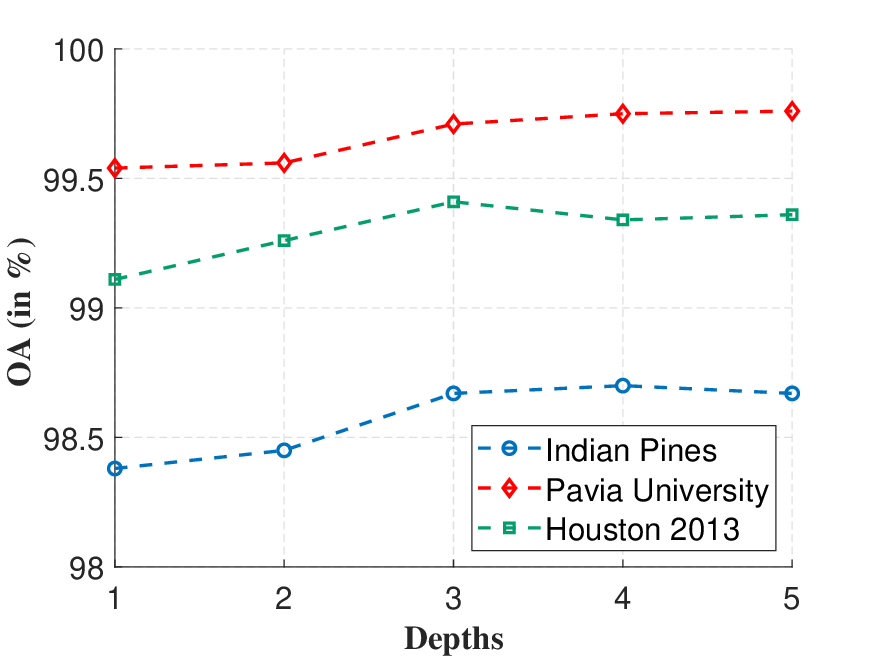}
		\caption{}
        \setlength{\belowdisplayskip}{43pt}
		\label{pseudo-I}
	\end{subfigure}
	\centering
	\begin{subfigure}{0.32\linewidth}
		\centering
		\includegraphics[width=5.5cm,height=4cm]{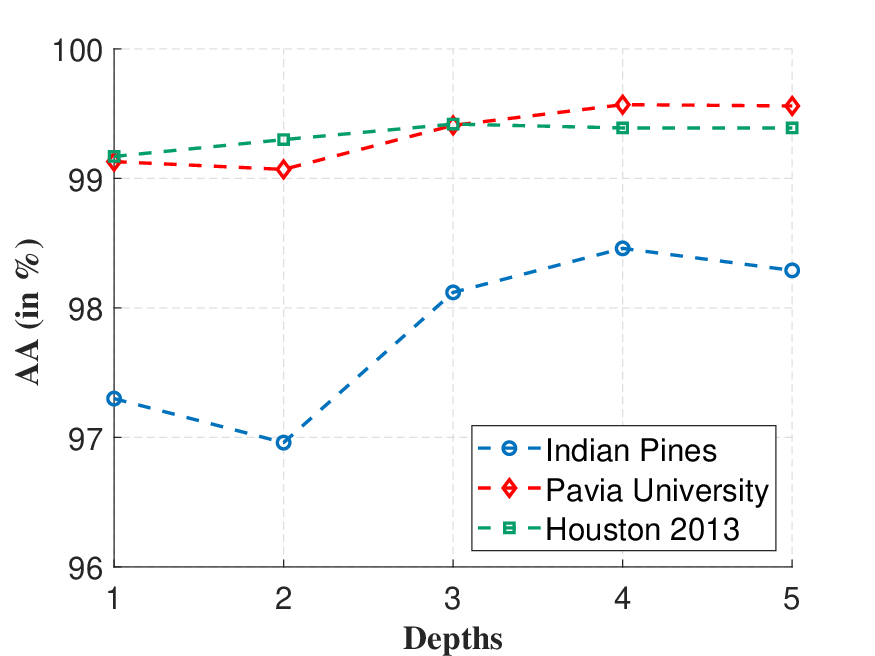}
		\caption{}
		\label{pseudo-I}
	\end{subfigure}
    \begin{subfigure}{0.3\linewidth}
		\centering
		\includegraphics[width=5.5cm,height=4cm]{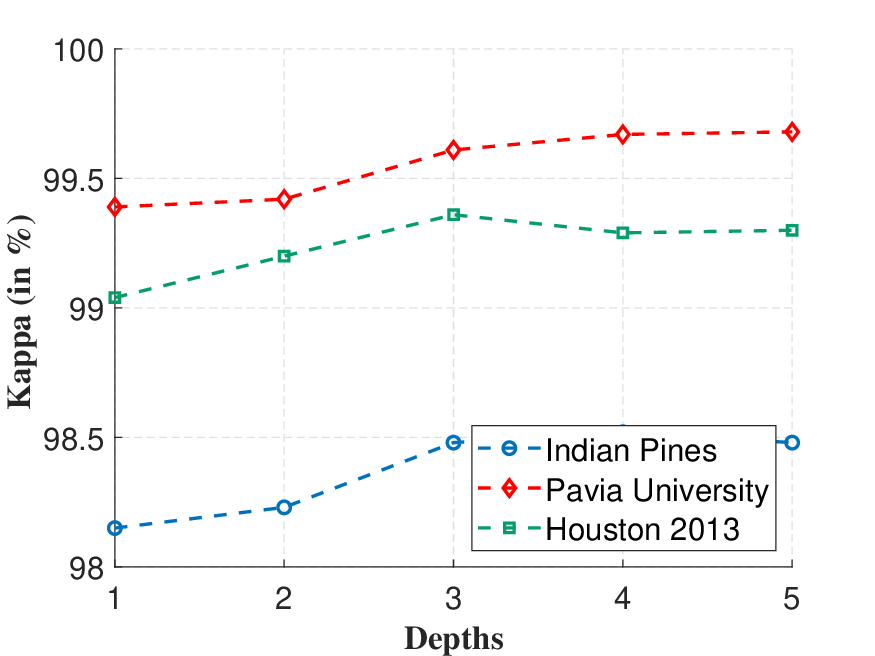}
		\caption{}
		\label{pseudo-I}
	\end{subfigure}
    \caption{Sensitivity analysis for the proposed method with different stage depths in terms of (a) OA, (b) AA, and (c) Kappa.}
	\label{fig:Depth}
\end{figure*}

To investigate the optimal structure of IGroupSS-Mamba for classification, a series of experiments are conducted by adjusting the feature embedding dimensions and the stage depths. Fig.~\ref{fig:Embed} and Fig.~\ref{fig:Depth} demonstrate the corresponding variations on three datasets in terms of OA (\%), AA (\%), and Kappa (\%). As can be observed, the performance exhibits a progressively increasing trend with the growth of embedding dimensions and stage depths. However, excessive high dimensions and deeper depths provide limited accuracy improvements but computational complexity and burdens. By trading off these metrics, the proposed IGroupSS-Mamba is constructed as a lightweight structure, with the embedding dimension set to 32 and the stage depth determined as 3.

\begin{figure*}[htbp]
	\centering
    \begin{subfigure}{0.32\linewidth}
		\centering
		\includegraphics[width=6cm,height=4cm]{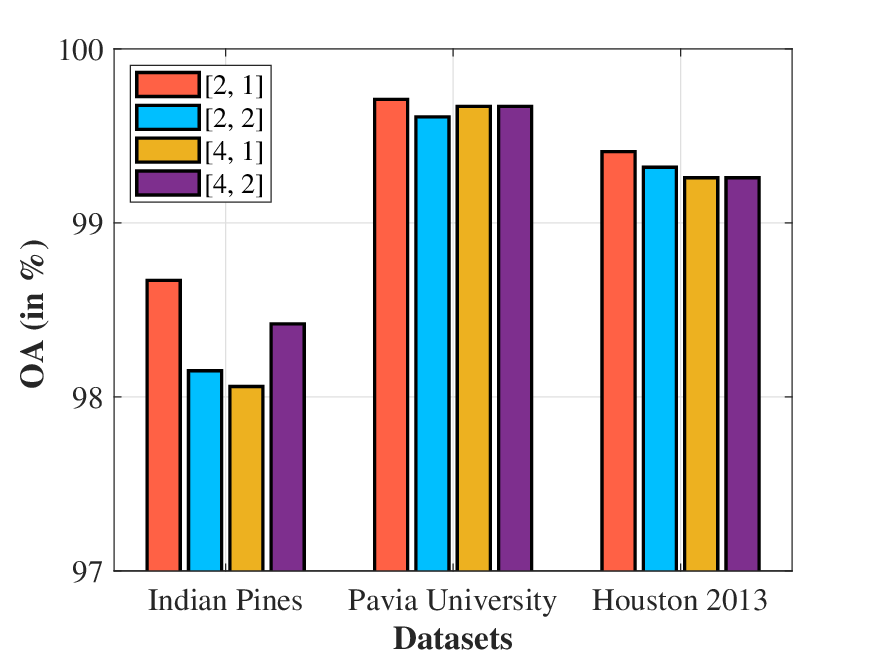}
		\caption{}
        \setlength{\belowdisplayskip}{43pt}
		\label{pseudo-I}
	\end{subfigure}
	\centering
	\begin{subfigure}{0.32\linewidth}
		\centering
		\includegraphics[width=6cm,height=4cm]{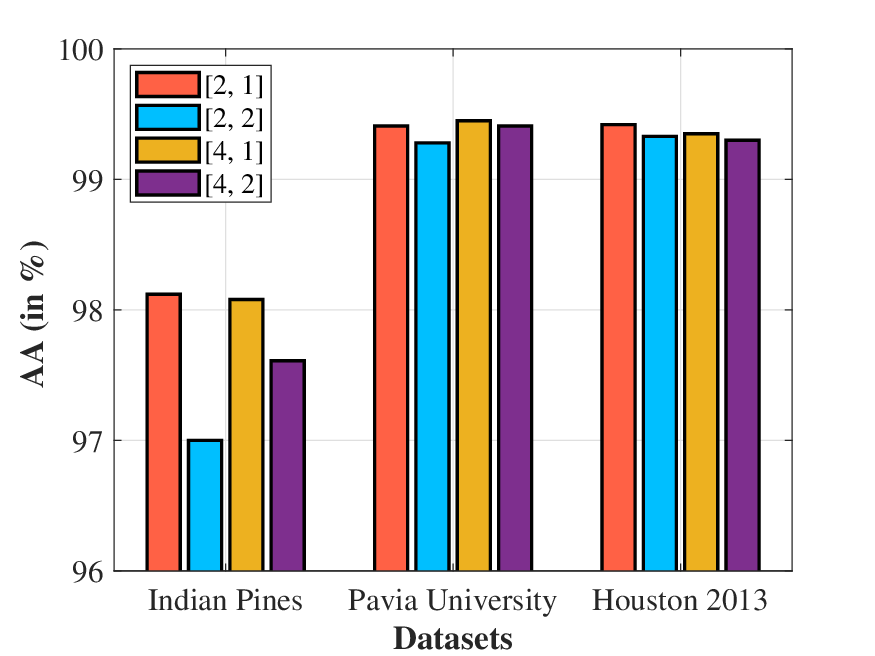}
		\caption{}
		\label{pseudo-I}
	\end{subfigure}
    \begin{subfigure}{0.3\linewidth}
		\centering
		\includegraphics[width=6cm,height=4cm]{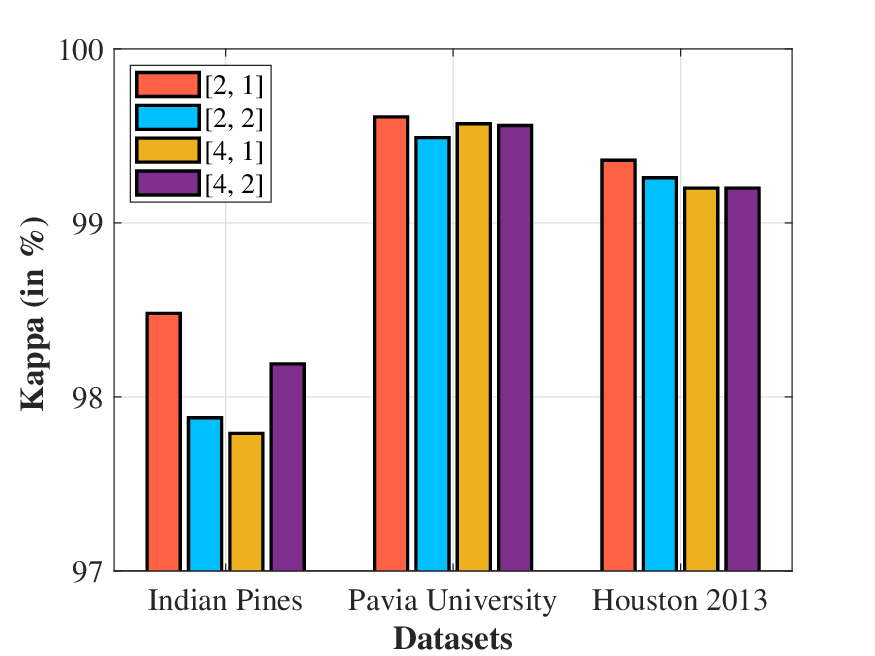}
		\caption{}
		\label{pseudo-I}
	\end{subfigure}
    \caption{Sensitivity analysis for the proposed method with different downsample scales in terms of (a) OA, (b) AA, and (c) Kappa.}
	\label{fig:Downsample}
\end{figure*}

\begin{table*}[!htb]
\centering
\caption{\textsc{Classification Accuracies of the Compared Methods in Terms of OA, AA, $\kappa$, and the Accuracies of Each CLasses for the Indian Pines Dataset. The Best Accuracies are Presented in Bold.}}
\setlength\tabcolsep{6.5pt}
\renewcommand\arraystretch{1.1}
\begin{tabular}{c|c|cc|cccccc|c}
\hline
\multirow{2}{*}{\textbf{Class}} & \multicolumn{1}{l|}{\textbf{Conventional}} & \multicolumn{2}{c|}{\textbf{CNN-based Methods}} & \multicolumn{6}{c|}{\textbf{Transformer-based Methods}}                              & \multirow{2}{*}{\textbf{\begin{tabular}[c]{@{}c@{}}IGroupSS\\ -Mamba\end{tabular}}} \\ \cline{2-10}
                                & SVM                                        & 2D-CNN                 & 3D-CNN                 & HSI-BERT & SF             & SSTN           & CASST & GSC-ViT        & SS-MTr         &                                                                                     \\ \hline
1                               & 20.73                                      & 76.09                  & 95.65                  & 48.78    & 100.00         & 76.07          & 78.26 & 53.66          & \textbf{100.0} & \textbf{100.0}                                                                      \\
2                               & 73.07                                      & 92.16                  & 88.31                  & 78.44    & 78.37          & 93.63          & 97.34 & \textbf{98.37} & 94.19          & 95.49                                                                               \\
3                               & 62.99                                      & 87.23                  & 84.94                  & 80.58    & 89.29          & 97.35          & 97.71 & 99.33          & 98.80          & \textbf{100.0}                                                                      \\
4                               & 50.70                                      & 64.98                  & 87.76                  & 51.17    & 81.22          & 78.06          & 99.58 & 97.18          & 92.41          & \textbf{98.12}                                                                      \\
5                               & 92.64                                      & 96.48                  & 96.48                  & 91.24    & 89.89          & 99.17          & 92.96 & \textbf{99.31} & 98.34          & 99.08                                                                               \\
6                               & 94.90                                      & 98.77                  & 96.16                  & 96.19    & 97.41          & 98.63          & 97.81 & \textbf{100.0} & \textbf{100.0} & 99.24                                                                               \\
7                               & 76.00                                      & 32.14                  & 78.57                  & 20.00    & 84.00          & 96.43          & 96.43 & \textbf{100.0} & \textbf{100.0} & \textbf{100.0}                                                                      \\
8                               & 96.63                                      & \textbf{100.0}         & \textbf{100.0}         & 98.83    & \textbf{100.0} & \textbf{100.0} & 99.79 & \textbf{100.0} & \textbf{100.0} & \textbf{100.0}                                                                      \\
9                               & 33.33                                      & 30.00                  & 70.00                  & 0.00     & 27.78          & 65.00          & 65.00 & 83.33          & 85.00          & \textbf{94.44}                                                                      \\
10                              & 68.69                                      & 91.87                  & 92.80                  & 78.37    & 94.97          & 97.84          & 92.80 & 98.97          & \textbf{99.59} & 98.29                                                                               \\
11                              & 85.16                                      & 95.93                  & 91.20                  & 92.85    & 93.44          & 98.04          & 98.37 & 98.37          & \textbf{99.59} & 99.55                                                                               \\
12                              & 64.89                                      & 93.42                  & 88.03                  & 62.47    & 81.65          & 88.53          & 91.40 & 96.25          & 94.10          & \textbf{97.38}                                                                      \\
13                              & 97.03                                      & 97.56                  & 94.15                  & 96.19    & 87.03          & 98.54          & 91.71 & \textbf{100.0} & 98.54          & 98.92                                                                               \\
14                              & 96.49                                      & 98.81                  & 99.05                  & 94.46    & 94.82          & 99.84          & 99.76 & 99.91          & 99.92          & \textbf{100.0}                                                                      \\
15                              & 55.33                                      & 81.35                  & 89.38                  & 87.89    & 96.83          & 99.48          & 95.34 & 90.78          & 93.01          & \textbf{100.0}                                                                      \\
16                              & 93.93                                      & 97.85                  & \textbf{100.0}         & 65.06    & 95.24          & 88.17          & 81.72 & 95.24          & 88.17          & 92.86                                                                               \\ \hline
OA (\%)                         & 79.82                                      & 93.34                  & 92.17                  & 85.45    & 90.67          & 96.56          & 96.65 & 98.28          & 97.92          & \textbf{98.71}                                                                      \\
AA (\%)                         & 72.03                                      & 83.41                  & 90.78                  & 71.41    & 87.00          & 92.18          & 92.25 & 94.42          & 96.35          & \textbf{98.33}                                                                      \\
Kappa                           & 76.84                                      & 92.39                  & 91.08                  & 83.36    & 89.36          & 96.07          & 96.18 & 98.04          & 97.63          & \textbf{98.53}                                                                      \\ \hline
\end{tabular}
\label{tab:IP}
\end{table*}

\begin{figure*}[!htb]
	\centering
	\includegraphics[width=0.85\textwidth]{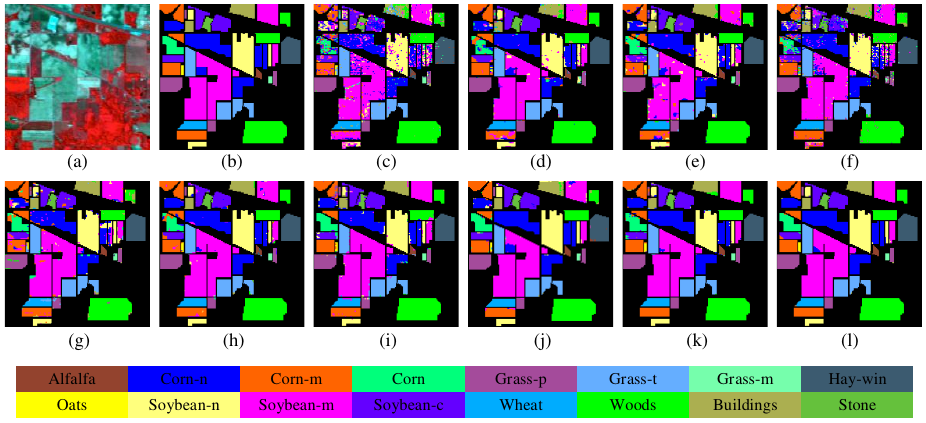}
	\caption{Classification maps obtained by the compared methods on the Indian Pines dataset. (a) Original data. (b) Reference map. (c) SVM. (d) 2D-CNN. (e) 3D-CNN. (f) HSI-BERT. (g) SF. (h) SSTN. (i) CASST. (j) GSC-ViT. (k) SS-MTr. (l) IGroupSS-Mamba.}
\label{fig:IP}
\end{figure*}

\subsubsection{Different Downsample Scales}
The downsampling operation is achieved by performing pixel aggregation along the spatial dimension. Fig.~\ref{fig:Downsample} investigates the impact of different aggregation scales on three datasets. Taking $\left[ {m,s} \right] = [2,1]$ as an example, the aggregation kernel is set to $2 \times 2$ and the aggregation stride is 1. From the results, different aggregation scales contribute variably to the extraction of multiscale spatial-spectral information, exhibiting slight performance fluctuations. In our experiments, the downsample scale $[2,1]$ is uniformly applied across all three datasets.

\subsection{Experimental Comparison With Competitive Approaches}
Table~\ref{tab:IP}-\ref{tab:HU13} summarizes the quantitative accuracies (OA (\%), AA (\%), and Kappa (\%)) on the Indian Pines, Pavia University, and Houston 2013 datasets, with the optimal results highlighted in bold. The corresponding visualization maps are presented in Fig.~\ref{fig:IP}-\ref{fig:HU13}.

\subsubsection{Indian Pines Dataset}
The experiments on the Indian Pines dataset are performed with 10\% of the reference samples. The quantitative classification accuracies are reported in Table~\ref{tab:IP}. As can be observed, the proposed IGroupSS-Mamba demonstrates significant superiority and effectiveness compared to other studied methods. Conventional SVM model and CNN-based methods including 2D-CNN and 3D-CNN are inherently constrained by manually crafted descriptors and local receptive fields, inevitably suffering from limited feature extraction capabilities in handling complex data. Transformer-based architectures establish long-range dependencies based on the attention mechanism. The classical SF method focuses on the groupwise spectral embedding to learn localized spectral representations, while ignoring the exploration of spatial contextual information. Compared to the proposed IGroupSS-Mamba, it demonstrates a substantial accuracy reduction of 8.04\% in OA, 11.33\% in AA, and 9.17\% in Kappa. Recent SS-MTr and GSC-ViT methods integrate the grouping strategy and multi-scale characteristics into the transformer network, respectively, which achieve slight performance advantages but remain inferior to IGroupSS-Mamba. Comparatively, IGroupSS-Mamba performs global spatial-spectral contextual modeling based on the Selective State Space Models. The interval-wise feature grouping strategy allows the model to leverage the complementarity of different scanning directions, and the cascading of spatial and spectral operators further exploits the correlation between spatial and spectral contextual, which significantly achieves the comprehensive extraction of multiscale spatial-spectral semantic representations. Compared to the suboptimal GSC-ViT approach, the quantitative improvements in terms of OA, AA, and Kappa up to 0.43\%, 3.91\%, and 0.49\%, respectively.

To further illustrate the differences in classification results, Fig.~\ref{fig:IP} presents the visualization maps of various comparison methods. It can be observed that SVM, 2D-CNN, 3D-CNN, and HSI-BERT generally suffer from significant noise interference and pronounced confusion in edge information. In contrast, the proposed IGroupSS-Mamba demonstrates the most consistent results with the ground truth, achieving clear category boundaries with minimal artifacts.

\subsubsection{Pavia University Dataset}
\begin{figure*}[!htb]
	\centering
	\includegraphics[width=0.85\textwidth]{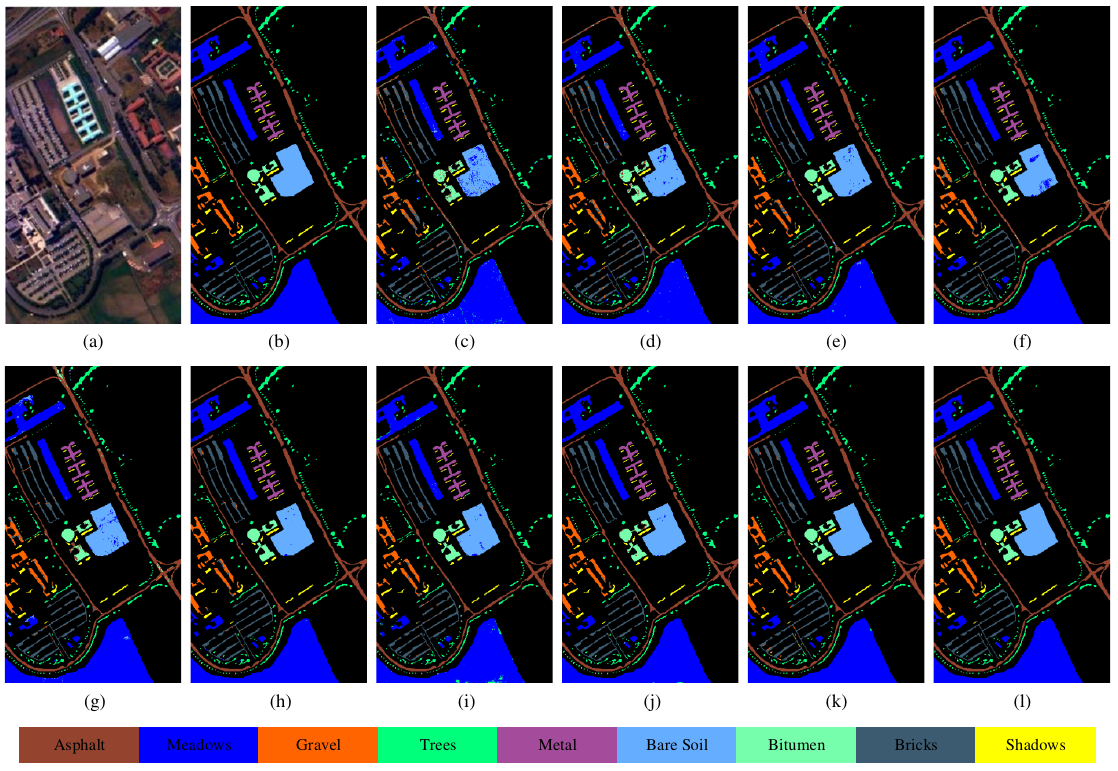}
	\caption{Classification maps obtained by the compared methods on the Pavia University dataset. (a) Original data. (b) Reference map. (c) SVM. (d) 2D-CNN. (e) 3D-CNN. (f) HSI-BERT. (g) SF. (h) SSTN. (i) CASST. (j) GSC-ViT. (k) SS-MTr. (l) IGroupSS-Mamba.}
\label{fig:PU}
\end{figure*}

The classification experiments on the Pavia University dataset are conducted with 5\% of the reference samples. Table~\ref{tab:PU} presents the quantitative comparison against various competitive approaches. As evidenced by the results, the proposed IGroupSS-Mamba achieves superior classification performance, particularly outperforming other methods in several categories such as Asphalt, Bitumen, and Self-blocking bricks. In contrast to the GSC-ViT method, IGroupSS-Mamba demonstrates notable improvements in OA, AA, and Kappa by 0.29\%, 0.36\%, and 0.38\%, respectively, further revealing the effectiveness of the grouping-based multidirectional scanning mechanism and multiscale extraction strategy. Additionally, the corresponding classification maps generated by comparison approaches are visualized in Fig.~\ref{fig:PU}. Due to the constrained feature extraction capability of hand-crafted descriptors, the conventional SVM method exhibits noticeable misclassification, particularly evident in the bare soil category. In contrast, the proposed IGroupSS-Mamba delivers the clearest and smoothest classifications across most regions.

\begin{table*}[!htb]
\centering
\caption{\textsc{Classification Accuracies of the Compared Methods in Terms of OA, AA, $\kappa$, and the Accuracies of Each CLasses for the Pavia University Dataset. The Best Accuracies are Presented in Bold.}}
\setlength\tabcolsep{6.5pt}
\renewcommand\arraystretch{1.1}
\begin{tabular}{c|c|cc|cccccc|c}
\hline
\multirow{2}{*}{\textbf{Class}} & \multicolumn{1}{l|}{\textbf{Conventional}} & \multicolumn{2}{c|}{\textbf{CNN-based Methods}} & \multicolumn{6}{c|}{\textbf{Transformer-based Methods}}                       & \multirow{2}{*}{\textbf{\begin{tabular}[c]{@{}c@{}}IGroupSS\\ -Mamba\end{tabular}}} \\ \cline{2-10}
                                & SVM                                        & 2D-CNN             & 3D-CNN                     & HSI-BERT & SF             & SSTN           & CASST & GSC-ViT & SS-MTr         &                                                                                     \\ \hline
1                               & 91.63                                      & 96.31              & 95.94                      & 97.63    & 94.75          & 99.52          & 99.17 & 99.63   & 99.32          & \textbf{99.86}                                                                      \\
2                               & 97.56                                      & 99.37              & 99.71                      & 99.93    & 98.14          & \textbf{99.98} & 99.95 & 99.90   & 99.90          & 99.94                                                                               \\
3                               & 73.90                                      & 82.22              & 89.76                      & 86.51    & 84.70          & 92.04          & 98.28 & 97.32   & 97.57          & \textbf{98.40}                                                                      \\
4                               & 92.10                                      & 94.35              & 97.36                      & 97.53    & 96.81          & \textbf{99.35} & 93.08 & 98.29   & 96.54          & 99.14                                                                               \\
5                               & 98.51                                      & 99.85              & \textbf{100.0}             & 99.61    & 99.22          & \textbf{100.0} & 98.66 & 99.89   & 99.78          & 99.61                                                                               \\
6                               & 86.04                                      & 93.65              & 97.41                      & 93.05    & 94.37          & 99.24          & 99.60 & 99.74   & \textbf{100.0} & \textbf{100.0}                                                                      \\
7                               & 84.09                                      & 89.17              & 96.69                      & 95.80    & 84.56          & 99.85          & 99.62 & 99.59   & 99.32          & \textbf{100.0}                                                                      \\
8                               & 90.25                                      & 88.73              & 93.92                      & 97.86    & 92.65          & 94.95          & 94.73 & 98.63   & 92.12          & \textbf{99.77}                                                                      \\
9                               & 99.22                                      & 97.57              & \textbf{100.0}             & 99.67    & \textbf{100.0} & 99.79          & 96.41 & 98.93   & 93.35          & 98.44                                                                               \\ \hline
OA (\%)                         & 92.75                                      & 95.77              & 97.62                      & 97.61    & 96.60          & 98.95          & 98.63 & 99.46   & 98.63          & \textbf{99.75}                                                                      \\
AA (\%)                         & 90.37                                      & 93.47              & 96.75                      & 96.40    & 93.91          & 98.30          & 97.72 & 99.10   & 97.55          & \textbf{99.46}                                                                      \\
Kappa                           & 90.35                                      & 94.37              & 96.84                      & 96.82    & 94.17          & 98.60          & 98.19 & 99.28   & 98.19          & \textbf{99.66}                                                                      \\ \hline
\end{tabular}
\label{tab:PU}
\end{table*}

\begin{table*}[!htb]
\centering
\caption{\textsc{Classification Accuracies of the Compared Methods in Terms of OA, AA, $\kappa$, and the Accuracies of Each CLasses for the Houston 2013 Dataset. The Best Accuracies are Presented in Bold.}}
\setlength\tabcolsep{6.5pt}
\renewcommand\arraystretch{1.1}
\begin{tabular}{c|c|cc|cccccc|c}
\hline
\multirow{2}{*}{\textbf{Class}} & \multicolumn{1}{l|}{\textbf{Conventional}} & \multicolumn{2}{c|}{\textbf{CNN-based Methods}} & \multicolumn{6}{c|}{\textbf{Transformer-based Methods}}                                       & \multirow{2}{*}{\textbf{\begin{tabular}[c]{@{}c@{}}IGroupSS\\ -Mamba\end{tabular}}} \\ \cline{2-10}
                                & SVM                                        & 2D-CNN                 & 3D-CNN                 & HIS-BERT & SF             & SSTN           & CASST          & GSC-ViT        & SS-MTr         &                                                                                     \\ \hline
1                               & 98.53                                      & 98.40                  & 97.04                  & 92.26    & 99.29          & 99.76          & 97.04          & 98.67          & 98.80          & \textbf{99.91}                                                                      \\
2                               & 98.05                                      & 99.20                  & 98.48                  & 83.15    & 98.05          & 99.68          & 98.01          & 99.73          & 99.60          & \textbf{99.82}                                                                      \\
3                               & 98.88                                      & \textbf{100.0}         & \textbf{100.0}         & 99.20    & \textbf{100.0} & \textbf{100.0} & 98.71          & 99.68          & \textbf{100.0} & 99.84                                                                               \\
4                               & 97.81                                      & 98.31                  & 98.95                  & 93.47    & 96.43          & \textbf{99.84} & 99.04          & 98.84          & 99.44          & 98.48                                                                               \\
5                               & 98.57                                      & 99.28                  & 99.44                  & 99.73    & \textbf{100.0} & \textbf{100.0} & \textbf{100.0} & \textbf{100.0} & \textbf{100.0} & \textbf{100.0}                                                                      \\
6                               & 92.64                                      & 92.62                  & \textbf{100.0}         & 78.76    & 95.89          & 96.00          & 95.38          & \textbf{100.0} & 98.15          & \textbf{100.0}                                                                      \\
7                               & 91.98                                      & 94.01                  & 93.38                  & 91.06    & 96.06          & 96.92          & 95.90          & 95.79          & \textbf{98.10} & 97.46                                                                               \\
8                               & 91.96                                      & 89.63                  & 94.29                  & 86.68    & 91.70          & 97.67          & 96.22          & 97.59          & 96.70          & \textbf{99.38}                                                                      \\
9                               & 85.09                                      & 83.87                  & 93.61                  & 89.60    & 94.59          & 97.36          & 95.53          & 91.66          & 98.80          & \textbf{99.73}                                                                      \\
10                              & 93.16                                      & 94.62                  & 96.33                  & 89.03    & 97.46          & 99.51          & 99.27          & 99.18          & 99.84          & \textbf{100.0}                                                                      \\
11                              & 87.01                                      & 90.53                  & 96.84                  & 95.31    & 90.56          & 99.67          & 97.89          & 99.01          & 98.79          & \textbf{100.0}                                                                      \\
12                              & 87.25                                      & 89.38                  & 95.46                  & 88.63    & 99.28          & \textbf{99.68} & 98.54          & 98.02          & 97.89          & 99.55                                                                               \\
13                              & 32.35                                      & 77.83                  & 94.24                  & 93.12    & 56.40          & 82.83          & 97.23          & 97.39          & 88.81          & \textbf{97.63}                                                                      \\
14                              & 98.91                                      & 99.30                  & 98.60                  & 97.92    & 98.44          & 99.53          & \textbf{100.0} & 99.74          & \textbf{100.0} & \textbf{100.0}                                                                      \\
15                              & 85.08                                      & \textbf{100.0}         & 99.85                  & 99.66    & 99.66          & \textbf{100.0} & 99.39          & \textbf{100.0} & \textbf{100.0} & \textbf{100.0}                                                                      \\ \hline
OA (\%)                         & 91.74                                      & 93.93                  & 96.77                  & 91.67    & 95.46          & 98.54          & 97.85          & 98.11          & 98.62          & \textbf{99.45}                                                                      \\
AA (\%)                         & 90.03                                      & 93.80                  & 97.10                  & 91.84    & 94.25          & 97.89          & 97.88          & 98.35          & 98.34          & \textbf{99.45}                                                                      \\
Kappa                           & 91.06                                      & 93.44                  & 96.51                  & 90.99    & 95.09          & 98.42          & 97.68          & 97.95          & 98.50          & \textbf{99.40}                                                                      \\ \hline
\end{tabular}
\label{tab:HU13}
\end{table*}

\begin{figure*}[!htb]
	\centering
	\includegraphics[width=0.92\textwidth]{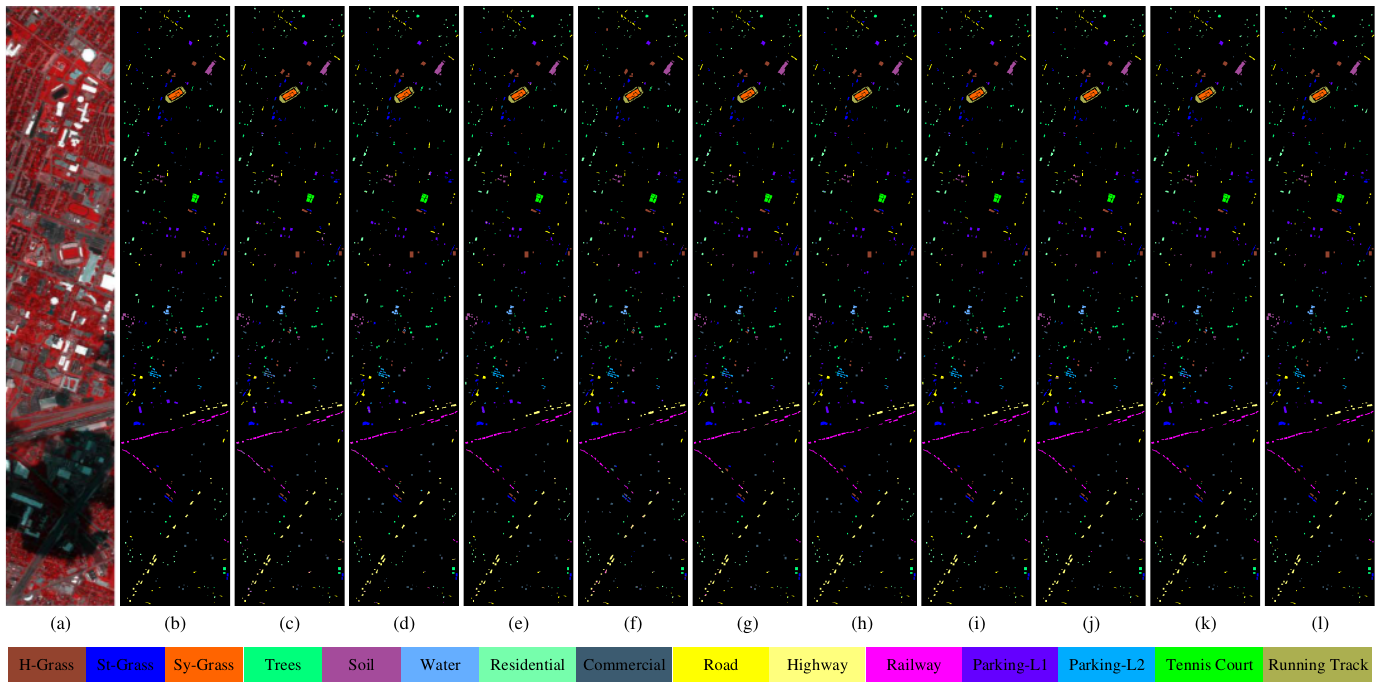}
	\caption{Classification maps obtained by the compared methods on the Houston 2013 dataset. (a) Original data. (b) Reference map. (c) SVM. (d) 2D-CNN. (e) 3D-CNN. (f) HSI-BERT. (g) SF. (h) SSTN. (i) CASST. (j) GSC-ViT. (k) SS-MTr. (l) IGroupSS-Mamba.}
\label{fig:HU13}
\end{figure*}

\subsubsection{Houston 2013 Dataset}
The experiments on the Houston 2013 dataset are executed with 10\% of the labeled samples. The classification results is illustrated in Table~\ref{tab:HU13}. As evident from the results, the proposed IGroupSS-Mamba consistently outperforms other competitive approaches by substantial margins, demonstrating the highest quantities across all three metrics. Restricted by the modeling capability, the suboptimal SS-MTr inevitably exhibits degraded classification consequences, with the reduction compared to IGroupSS-Mamba reaching 0.83\% for OA, 1.11\% for AA, and 0.9\% for Kappa, respectively. According to the visualization maps in Fig.~\ref{fig:HU13}, IGroupSS-Mamba provides the most precise prediction details, despite the predominantly discrete and localized sample targets in this scenario. These significant improvements further validate the potential of State Space Models in HSI classification.

\subsection{Ablation Study}
\subsubsection{Effectiveness of Interval Grouping Strategy}
\begin{figure*}[htbp]
	\centering
	\begin{subfigure}{0.32\linewidth}
		\centering
		\includegraphics[width=6cm,height=4cm]{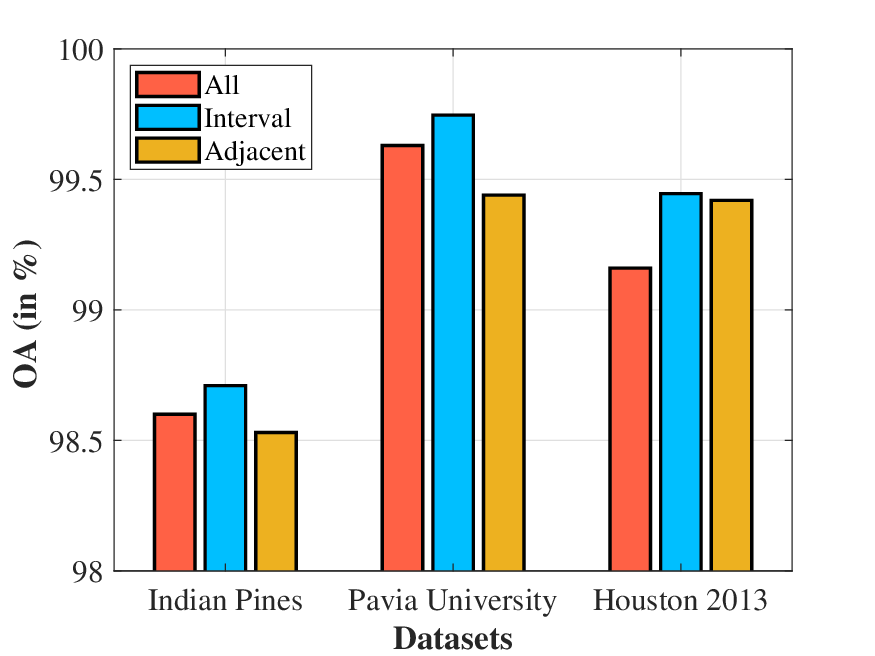}
		\caption{}
        \setlength{\belowdisplayskip}{43pt}
		\label{pseudo-I}
	\end{subfigure}
	\centering
	\begin{subfigure}{0.32\linewidth}
		\centering
		\includegraphics[width=6cm,height=4cm]{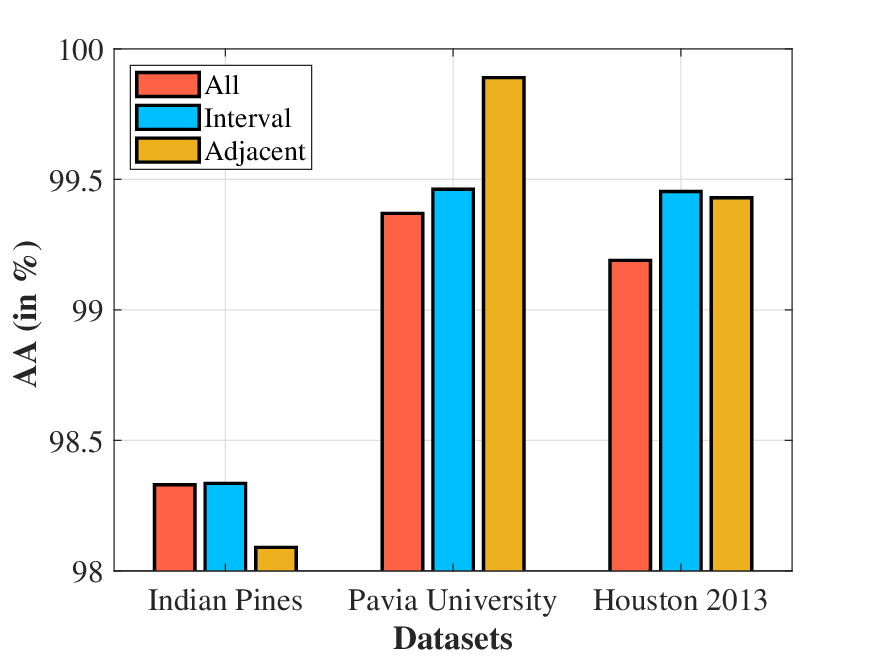}
		\caption{}
		\label{pseudo-I}
	\end{subfigure}
    \begin{subfigure}{0.3\linewidth}
		\centering
		\includegraphics[width=6cm,height=4cm]{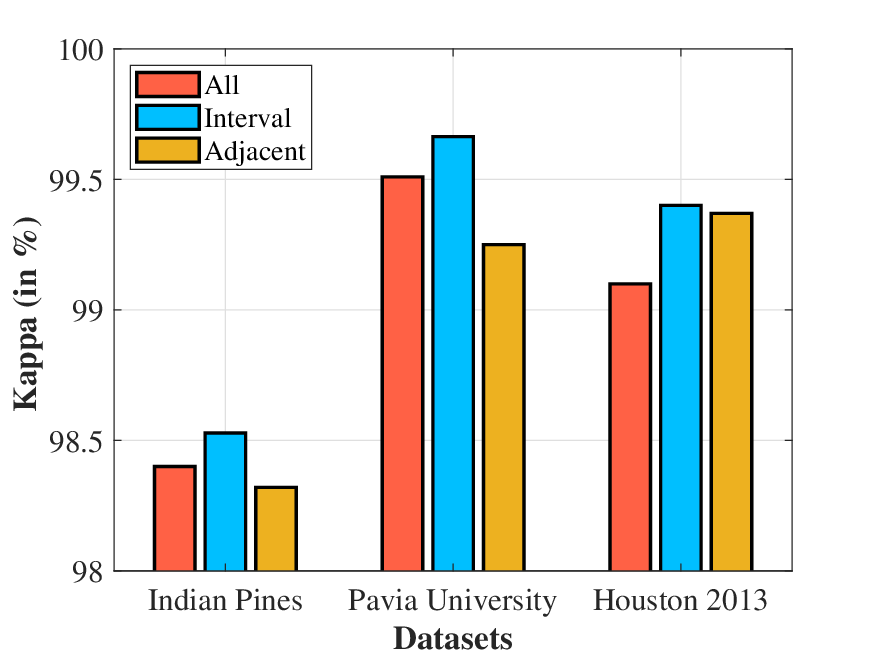}
		\caption{}
		\label{pseudo-I}
	\end{subfigure}
    \caption{Effectiveness analysis of different grouping strategies on three datasets in terms of (a) OA, (b) AA, and (c) Kappa.}
	\label{fig:Group}
\end{figure*}

Fig.~\ref{fig:Group} illustrates the performance variations with different grouping strategies on three HSI datasets. The All denotes applying multidirectional scanning into all spectral bands without grouping. The Adjacent denotes performing feature grouping in an adjacent manner, generating four local HSI cubes with continuous features. The Interval refers to the proposed interval grouping strategy as Eq.~(\ref{eq:Interval}), constructing the non-overlapping feature groups in an interval manner. As can be observed, the proposed Interval strategy demonstrates more competitive superiority and advantages in HSI classification, while the All and Adjacent schemes exhibit slight performance degradations and fluctuations. The potential reason might be that the All strategy encounters challenges with information redundancy in sequence modeling, which arises from the excessively high spectral dimensionality. The Adjacent scheme tends to capture local information of the spectrum, as depicted in Fig.~\ref{fig:Correlation}(b), thereby constraining its ability to adequately explore global spatial-spectral correlations. Comparatively, the Interval strategy accounts for the similarity between adjacent spectral bands, allowing each feature group to comprehensively preserve global characteristics without redundancy, as illustrated in Fig.~\ref{fig:Correlation}(c). These notable advantages significantly reveal the feasibility and superiority of the proposed interval grouping strategy.

\subsubsection{Effectiveness of Spatial and Spectral Operators}
To highlight the specific effectiveness of the spatial and spectral operators in IGSSB, ablation experiments are conducted on three HSI datasets by removing either the spectral or spatial into a single-operator setup. The classification results in terms of accuracy metrics are illustrated in Table~\ref{tab:Operator}. It can be observed that the individual Spatial Operator and Spectral Operator each achieve significant classification performance, revealing the effectiveness of the proposed scanning directions along the spatial and spectral dimensions. Additionally, the cascade of spatial and spectral operators further enhances the classification capability, demonstrating competitive advantages and strong robustness across all datasets. Specifically, the Cascaded Spa-Spe surpasses the Spatial Operator by an average of 0.09\% in OA, 0.28\% in AA, and 0.11\% in Kappa. It exceeds the Spectral Operator by margins of 0.14\%, 0.26\%, and 0.14\% for OA, AA, and Kappa, respectively. These disparities significantly accentuate the superiority of the cascading structure, which adequately takes advantage of the complementarity of spatial and spectral information of hyperspectral data.

\begin{table*}[!htb]
\centering
\caption{\textsc{Effectiveness analysis of Spatial and Spectral Operators on the Indian Pines, Pavia University, and Houston 2013 Datasets.}}
\setlength\tabcolsep{7.5pt}
\renewcommand\arraystretch{1.1}
\begin{tabular}{c|ccccccccc}
\hline
\multirow{2}{*}{} & \multicolumn{3}{c}{\textbf{Indian Pines}}                             & \multicolumn{3}{c}{\textbf{Pavia University}}                         & \multicolumn{3}{c}{\textbf{Houston2013}}         \\ \cline{2-10}
                  & OA (\%)        & AA (\%)        & \multicolumn{1}{c|}{Kappa (\%)}     & OA (\%)        & AA (\%)        & \multicolumn{1}{c|}{Kappa (\%)}     & OA (\%)        & AA (\%)        & Kappa (\%)     \\ \hline
Spatial Operator          & 98.56          & 97.69          & \multicolumn{1}{c|}{98.35}          & 99.66          & 99.29          & \multicolumn{1}{c|}{99.55}          & 99.41          & 99.43          & 99.36          \\
Spectral Operator         & 98.42          & 97.84          & \multicolumn{1}{c|}{98.20}          & 99.65          & 99.25          & \multicolumn{1}{c|}{99.53}          & 99.35          & 99.38          & 99.30          \\
\textbf{Cascaded Spa-Spe}       & \textbf{98.71} & \textbf{98.33} & \multicolumn{1}{c|}{\textbf{98.53}} & \textbf{99.75} & \textbf{99.46} & \multicolumn{1}{c|}{\textbf{99.66}} & \textbf{99.45} & \textbf{99.45} & \textbf{99.40} \\ \hline
\end{tabular}
\label{tab:Operator}
\end{table*}

\begin{table*}[!htb]
\centering
\caption{\textsc{Computational parameter analysis of various comparison approaches in terms of parameters, Flops, and inference time on the PU dataset.}}
\setlength\tabcolsep{5.1pt}
\renewcommand\arraystretch{1.1}
\begin{tabular}{c|c|ccccccccc}
\hline
                                           & Metrics            & 2D-CNN & 3D-CNN & HIS-BERT & SF     & SSTN   & CASST  & GSC-ViT & SS-MTr & \begin{tabular}[c]{@{}c@{}}IGroupSS\\ -Mamba\end{tabular} \\ \hline
\multirow{3}{*}{\textbf{Indian Pines}}     & Params (M)         & 0.0173 & 0.0165 & 0.3951   & 0.1226 & 0.3615 & 6.5275 & 0.5636  & 1.6153 & 0.0575                                                    \\
                                           & Flops (G)          & 0.0003 & 0.0024 & 7.1523   & 0.0245 & 0.0033 & 0.9893 & 0.0201  & 0.1199 & 0.0095                                                    \\
                                           & Inference time (s) & 0.193  & 0.27   & 9.021    & 2.139  & 0.554  & 3.23   & 1.64    & 1.36   & 4.79                                                      \\ \hline
\multirow{3}{*}{\textbf{Pavia University}} & Params (M)         & 0.0101 & 0.0073 & 0.3006   & 0.1222 & 0.3577 & 6.5275 & 0.0778  & 1.4622 & 0.0573                                                    \\
                                           & Flops (G)          & 0.0001 & 0.0012 & 5.441    & 0.0126 & 0.0032 & 0.9893 & 0.0049  & 0.1197 & 0.0095                                                    \\
                                           & Inference time (s) & 0.56   & 0.702  & 36.13    & 6.55   & 2.18   & 13.4   & 13.256  & 5.29   & 19.76                                                     \\ \hline
\multirow{3}{*}{\textbf{Houston 2013}}     & Params (M)         & 0.0132 & 0.0123 & 0.3773   & 0.1133 & 0.3609 & 6.5316 & 0.0887  & 1.3667 & 0.0575                                                    \\
                                           & Flops (G)          & 0.0002 & 0.0017 & 6.8135   & 0.0163 & 0.0032 & 0.9893 & 0.0056  & 0.0377 & 0.0095                                                    \\
                                           & Inference time (s) & 0.247  & 0.313  & 12.218   & 1.773  & 0.78   & 161.34 & 45.41   & 1.11   & 3.30                                                      \\ \hline
\end{tabular}
\label{tab:Complex}
\end{table*}

\subsection{Analysis of Computational Complexity}
As illustrated in Table~\ref{tab:Complex}, the computational complexity of the proposed IGroupSS-Mamba on all datasets is detailed, focusing on model parameters, Flops, and inference time. Parameters represent the number of parameters in the entire network. Flops signifies the number of floating-point operations performed by the model. Although CNN-based methods exhibit relatively low computational burdens, their ability to capture long-range dependencies is limited by local receptive fields. Transformer-based architectures generally suffer from higher resource consumption due to the inherent multi-head self-attention (MHSA) modules. Benefiting from the linear sequential modeling mechanism of SSMs and feature grouping strategy, the proposed IGroupSS-Mamba consistently demonstrates reduced computational parameters and Flops across various datasets. The inference time is also comparatively acceptable compared to other approaches. In summary, IGroupSS-Mamba presents a competitive balance between computational efficiency and classification performance, underscoring significant potentiality and viability for HSI classification.

\section{Conclusion}
\label{sec:conc}
In this paper, we introduce an Interval Group Spatial-Spectral Mamba framework (IGroupSS-Mamba), which supports multidirectional and multi-scale spatial-spectral information processing in a grouping and hierarchical manner. Benefiting from the Interval Group S6 Mechanis (IGSM), IGroupSS-Mamba achieves non-redundant spatial-spectral dependencies modeling across different feature groups, which enjoys the complementarity of different scanning directions while alleviating information interference. Meanwhile, the design of Interval Group Spatial-Spectral Block (IGSSB) enables IGroupSS-Mamba to effectively leverage the complementary strengths of spectral and spatial information inherent in hyperspectral data. Extensive experiments indicate that IGroupSS-Mamba efficiently overcomes the performance and efficiency limitations in state-of-the-art CNN-based and Transformer-based HSI architectures, which provides a feasible solution for HSI classification task. Future research will focus on integrating the Mamba model with alternative network architectures and extending its application to more challenging hyperspectral scenarios.

\bibliographystyle{IEEEtran_doi}
\bibliography{deepcs}

\begin{thebibliography}{10}
\providecommand{\url}[1]{#1}
\csname url@samestyle\endcsname
\providecommand{\newblock}{\relax}
\providecommand{\bibinfo}[2]{#2}
\providecommand{\BIBentrySTDinterwordspacing}{\spaceskip=0pt\relax}
\providecommand{\BIBentryALTinterwordstretchfactor}{4}
\providecommand{\BIBentryALTinterwordspacing}{\spaceskip=\fontdimen2\font plus
\BIBentryALTinterwordstretchfactor\fontdimen3\font minus \fontdimen4\font\relax}
\providecommand{\BIBforeignlanguage}[2]{{%
\expandafter\ifx\csname l@#1\endcsname\relax
\typeout{** WARNING: IEEEtran.bst: No hyphenation pattern has been}%
\typeout{** loaded for the language `#1'. Using the pattern for}%
\typeout{** the default language instead.}%
\else
\language=\csname l@#1\endcsname
\fi
#2}}
\providecommand{\BIBdecl}{\relax}
\BIBdecl

\bibitem{ni2020mineral}
L.~Ni, H.~Xu, and X.~Zhou, ``Mineral identification and mapping by synthesis of hyperspectral vnir/swir and multispectral tir remotely sensed data with different classifiers,'' \emph{IEEE Journal of Selected Topics in Applied Earth Observations and Remote Sensing}, vol.~13, pp. 3155--3163, 2020.

\bibitem{siebels2020estimation}
K.~Siebels, K.~Go{\"\i}ta, and M.~Germain, ``Estimation of mineral abundance from hyperspectral data using a new supervised neighbor-band ratio unmixing approach,'' \emph{IEEE Transactions on Geoscience and Remote Sensing}, vol.~58, no.~10, pp. 6754--6766, 2020.

\bibitem{ardouin2007demonstration}
J.-P. Ardouin, J.~L{\'e}vesque, and T.~A. Rea, ``A demonstration of hyperspectral image exploitation for military applications,'' in \emph{2007 10th International Conference on Information Fusion}.\hskip 1em plus 0.5em minus 0.4em\relax IEEE, 2007, pp. 1--8.

\bibitem{peyghambari2021hyperspectral}
S.~Peyghambari and Y.~Zhang, ``Hyperspectral remote sensing in lithological mapping, mineral exploration, and environmental geology: an updated review,'' \emph{Journal of Applied Remote Sensing}, vol.~15, no.~3, pp. 031\,501--031\,501, 2021.

\bibitem{lu2020recent}
B.~Lu, P.~D. Dao, J.~Liu, Y.~He, and J.~Shang, ``Recent advances of hyperspectral imaging technology and applications in agriculture,'' \emph{Remote Sensing}, vol.~12, no.~16, p. 2659, 2020.

\bibitem{tu2024ncglf2}
B.~Tu, Q.~Ren, J.~Li, Z.~Cao, Y.~Chen, and A.~Plaza, ``Ncglf2: Network combining global and local features for fusion of multisource remote sensing data,'' \emph{Information Fusion}, vol. 104, p. 102192, 2024.

\bibitem{melgani2004classification}
F.~Melgani and L.~Bruzzone, ``Classification of hyperspectral remote sensing images with support vector machines,'' \emph{IEEE Transactions on geoscience and remote sensing}, vol.~42, no.~8, pp. 1778--1790, 2004.

\bibitem{camps2013advances}
G.~Camps-Valls, D.~Tuia, L.~Bruzzone, and J.~A. Benediktsson, ``Advances in hyperspectral image classification: Earth monitoring with statistical learning methods,'' \emph{IEEE signal processing magazine}, vol.~31, no.~1, pp. 45--54, 2013.

\bibitem{huang2019dimensionality}
H.~Huang, G.~Shi, H.~He, Y.~Duan, and F.~Luo, ``Dimensionality reduction of hyperspectral imagery based on spatial--spectral manifold learning,'' \emph{IEEE transactions on cybernetics}, vol.~50, no.~6, pp. 2604--2616, 2019.

\bibitem{lunga2013manifold}
D.~Lunga, S.~Prasad, M.~M. Crawford, and O.~Ersoy, ``Manifold-learning-based feature extraction for classification of hyperspectral data: A review of advances in manifold learning,'' \emph{IEEE Signal Processing Magazine}, vol.~31, no.~1, pp. 55--66, 2013.

\bibitem{fauvel2008spectral}
M.~Fauvel, J.~A. Benediktsson, J.~Chanussot, and J.~R. Sveinsson, ``Spectral and spatial classification of hyperspectral data using svms and morphological profiles,'' \emph{IEEE Transactions on Geoscience and Remote Sensing}, vol.~46, no.~11, pp. 3804--3814, 2008.

\bibitem{dalla2010extended}
M.~Dalla~Mura, J.~Atli~Benediktsson, B.~Waske, and L.~Bruzzone, ``Extended profiles with morphological attribute filters for the analysis of hyperspectral data,'' \emph{International Journal of Remote Sensing}, vol.~31, no.~22, pp. 5975--5991, 2010.

\bibitem{duan2021semisupervised}
Y.~Duan, H.~Huang, and T.~Wang, ``Semisupervised feature extraction of hyperspectral image using nonlinear geodesic sparse hypergraphs,'' \emph{IEEE Transactions on Geoscience and Remote Sensing}, vol.~60, pp. 1--15, 2021.

\bibitem{hu2015deep}
W.~Hu, Y.~Huang, L.~Wei, F.~Zhang, and H.~Li, ``Deep convolutional neural networks for hyperspectral image classification,'' \emph{Journal of Sensors}, vol. 2015, pp. 1--12, 2015.

\bibitem{zhong2017spectral}
Z.~Zhong, J.~Li, Z.~Luo, and M.~Chapman, ``Spectral--spatial residual network for hyperspectral image classification: A 3-d deep learning framework,'' \emph{IEEE Transactions on Geoscience and Remote Sensing}, vol.~56, no.~2, pp. 847--858, 2018.

\bibitem{he2019hsi}
J.~He, L.~Zhao, H.~Yang, M.~Zhang, and W.~Li, ``Hsi-bert: Hyperspectral image classification using the bidirectional encoder representation from transformers,'' \emph{IEEE Transactions on Geoscience and Remote Sensing}, vol.~58, no.~1, pp. 165--178, 2019.

\bibitem{hong2022spectralformer}
D.~Hong, Z.~Han, J.~Yao, L.~Gao, B.~Zhang, A.~Plaza, and J.~Chanussot, ``Spectralformer: Rethinking hyperspectral image classification with transformers,'' \emph{IEEE Transactions on Geoscience and Remote Sensing}, vol.~60, pp. 1--15, 2022.

\bibitem{zhao2024hyperspectral}
Z.~Zhao, X.~Xu, S.~Li, and A.~Plaza, ``Hyperspectral image classification using groupwise separable convolutional vision transformer network,'' \emph{IEEE Transactions on Geoscience and Remote Sensing}, 2024.

\bibitem{gu2023mamba}
A.~Gu and T.~Dao, ``Mamba: Linear-time sequence modeling with selective state spaces,'' \emph{arXiv preprint arXiv:2312.00752}, 2023.

\bibitem{liu2024vmamba}
Y.~Liu, Y.~Tian, Y.~Zhao, H.~Yu, L.~Xie, Y.~Wang, Q.~Ye, and Y.~Liu, ``Vmamba: Visual state space model,'' \emph{arXiv preprint arXiv:2401.10166}, 2024.

\bibitem{lee2016contextual}
H.~Lee and H.~Kwon, ``Contextual deep cnn based hyperspectral classification,'' in \emph{2016 IEEE international geoscience and remote sensing symposium (IGARSS)}.\hskip 1em plus 0.5em minus 0.4em\relax IEEE, 2016, pp. 3322--3325.

\bibitem{hamida20183}
A.~B. Hamida, A.~Benoit, P.~Lambert, and C.~B. Amar, ``3-d deep learning approach for remote sensing image classification,'' \emph{IEEE Transactions on geoscience and remote sensing}, vol.~56, no.~8, pp. 4420--4434, 2018.

\bibitem{roy2020hybridsn}
S.~K. Roy, G.~Krishna, S.~R. Dubey, and B.~B. Chaudhuri, ``Hybridsn: Exploring 3-d--2-d cnn feature hierarchy for hyperspectral image classification,'' \emph{IEEE Geoscience and Remote Sensing Letters}, vol.~17, no.~2, pp. 277--281, 2020.

\bibitem{yang2022hyperspectral}
X.~Yang, W.~Cao, Y.~Lu, and Y.~Zhou, ``Hyperspectral image transformer classification networks,'' \emph{IEEE Transactions on Geoscience and Remote Sensing}, vol.~60, pp. 1--15, 2022.

\bibitem{roy2023spectral}
S.~K. Roy, A.~Deria, C.~Shah, J.~M. Haut, Q.~Du, and A.~Plaza, ``Spectral--spatial morphological attention transformer for hyperspectral image classification,'' \emph{IEEE Transactions on Geoscience and Remote Sensing}, vol.~61, pp. 1--15, 2023.

\bibitem{zhang2022convolution}
J.~Zhang, Z.~Meng, F.~Zhao, H.~Liu, and Z.~Chang, ``Convolution transformer mixer for hyperspectral image classification,'' \emph{IEEE Geoscience and Remote Sensing Letters}, vol.~19, pp. 1--5, 2022.

\bibitem{10038735}
E.~Ouyang, B.~Li, W.~Hu, G.~Zhang, L.~Zhao, and J.~Wu, ``\href{http://dx.doi.org/10.1109/TGRS.2023.3242978}{When multigranularity meets spatial–spectral attention: A hybrid transformer for hyperspectral image classification},'' \emph{IEEE Transactions on Geoscience and Remote Sensing}, vol.~61, pp. 1--18, 2023.

\bibitem{zhong2022spectral}
Z.~Zhong, Y.~Li, L.~Ma, J.~Li, and W.-S. Zheng, ``Spectral--spatial transformer network for hyperspectral image classification: A factorized architecture search framework,'' \emph{IEEE Transactions on Geoscience and Remote Sensing}, vol.~60, pp. 1--15, 2022.

\bibitem{peng2022spatial}
Y.~Peng, Y.~Zhang, B.~Tu, Q.~Li, and W.~Li, ``Spatial--spectral transformer with cross-attention for hyperspectral image classification,'' \emph{IEEE Transactions on Geoscience and Remote Sensing}, vol.~60, pp. 1--15, 2022.

\bibitem{mei2022hyperspectral}
S.~Mei, C.~Song, M.~Ma, and F.~Xu, ``Hyperspectral image classification using group-aware hierarchical transformer,'' \emph{IEEE Transactions on Geoscience and Remote Sensing}, vol.~60, pp. 1--14, 2022.

\bibitem{kalman1960new}
R.~E. Kalman, ``A new approach to linear filtering and prediction problems,'' 1960.

\bibitem{gu2022efficiently}
A.~Gu, K.~Goel, and C.~R{\'e}, ``Efficiently modeling long sequences with structured state spaces,'' \emph{arXiv preprint arXiv:2111.00396}, 2022.

\bibitem{zhu2024vision}
L.~Zhu, B.~Liao, Q.~Zhang, X.~Wang, W.~Liu, and X.~Wang, ``Vision mamba: Efficient visual representation learning with bidirectional state space model,'' \emph{arXiv preprint arXiv:2401.09417}, 2024.

\bibitem{huang2024localmamba}
T.~Huang, X.~Pei, S.~You, F.~Wang, C.~Qian, and C.~Xu, ``Localmamba: Visual state space model with windowed selective scan,'' \emph{arXiv preprint arXiv:2403.09338}, 2024.

\bibitem{chen2024changemamba}
H.~Chen, J.~Song, C.~Han, J.~Xia, and N.~Yokoya, ``Changemamba: Remote sensing change detection with spatio-temporal state space model,'' \emph{arXiv preprint arXiv:2404.03425}, 2024.

\bibitem{zhao2024rs}
S.~Zhao, H.~Chen, X.~Zhang, P.~Xiao, L.~Bai, and W.~Ouyang, ``Rs-mamba for large remote sensing image dense prediction,'' \emph{arXiv preprint arXiv:2404.02668}, 2024.

\bibitem{ma2024rs}
X.~Ma, X.~Zhang, and M.-O. Pun, ``Rs 3 mamba: Visual state space model for remote sensing image semantic segmentation,'' \emph{IEEE Geoscience and Remote Sensing Letters}, 2024.

\bibitem{zhu2024samba}
Q.~Zhu, Y.~Cai, Y.~Fang, Y.~Yang, C.~Chen, L.~Fan, and A.~Nguyen, ``Samba: Semantic segmentation of remotely sensed images with state space model,'' \emph{arXiv preprint arXiv:2404.01705}, 2024.

\bibitem{renard2008denoising}
N.~Renard, S.~Bourennane, and J.~Blanc-Talon, ``Denoising and dimensionality reduction using multilinear tools for hyperspectral images,'' \emph{IEEE Geoscience and Remote Sensing Letters}, vol.~5, no.~2, pp. 138--142, 2008.

\bibitem{ma2013hughes}
W.~Ma, C.~Gong, Y.~Hu, P.~Meng, and F.~Xu, ``The hughes phenomenon in hyperspectral classification based on the ground spectrum of grasslands in the region around qinghai lake,'' in \emph{International Symposium on Photoelectronic Detection and Imaging 2013: Imaging Spectrometer Technologies and Applications}, vol. 8910.\hskip 1em plus 0.5em minus 0.4em\relax SPIE, 2013, pp. 363--373.

\bibitem{elfwing2018sigmoid}
S.~Elfwing, E.~Uchibe, and K.~Doya, ``Sigmoid-weighted linear units for neural network function approximation in reinforcement learning,'' \emph{Neural networks}, vol. 107, pp. 3--11, 2018.

\bibitem{huang2023spectral}
L.~Huang, Y.~Chen, and X.~He, ``Spectral-spatial masked transformer with supervised and contrastive learning for hyperspectral image classification,'' \emph{IEEE Transactions on Geoscience and Remote Sensing}, 2023.

\end{thebibliography}
\renewenvironment{IEEEbiography}[1] {\IEEEbiographynophoto{#1}}{\endIEEEbiographynophoto}

\end{document}